\RequirePackage{fix-cm}
\documentclass[smallextended]{svjour3}       
\smartqed  
\usepackage{graphicx}

\usepackage{booktabs} 
\usepackage{multirow}
\usepackage{color}
\usepackage{amsmath}
\usepackage{longtable}
\newcommand{\tabincell}[2]{\begin{tabular}{@{}#1@{}}#2\end{tabular}}
\usepackage[ruled,vlined]{algorithm2e}
\usepackage{algorithmic} 

\usepackage[misc]{ifsym}
\begin{document}

\title{Copy-move Forgery Detection based on Convolutional Kernel Network
}


\author{Yaqi Liu $^{1,2}$        \and
        Qingxiao Guan $^{1,2}$   \and
       Xianfeng Zhao $^{1,2}$
}


\institute{
\Letter \hspace{0.1cm} Xianfeng Zhao \\
\email{zhaoxianfeng@iie.ac.cn}\\
1. State Key Laboratory of Information Security, Institute of Information Engineering, Chinese Academy of Sciences, Beijing 100093, China \\
2. School of Cyber Security, University of Chinese Academy of Sciences, Beijing 100093, China
}

\date{Received: date / Accepted: date}

\maketitle

\begin{abstract}
In this paper, a copy-move forgery detection method based on Convolutional Kernel Network is proposed. Different from methods based on conventional hand-crafted features, Convolutional Kernel Network is a kind of data-driven local descriptor with the deep convolutional structure. Thanks to the development of deep learning theories and widely available datasets, the data-driven methods can achieve competitive performance on different conditions for its excellent discriminative capability. Besides, our Convolutional Kernel Network is reformulated as a series of matrix computations and convolutional operations which are easy to parallelize and accelerate by GPU, leading to high efficiency. Then, appropriate preprocessing and postprocessing for Convolutional Kernel Network are adopted to achieve copy-move forgery detection. Particularly, a segmentation-based keypoints distribution strategy is proposed and a GPU-based adaptive oversegmentation method is adopted. Numerous experiments are conducted to demonstrate the effectiveness and robustness of the GPU version of Convolutional Kernel Network, and the state-of-the-art performance of the proposed copy-move forgery detection method based on Convolutional Kernel Network.
\end{abstract}

\section{Introduction}
\label{sect:introduction}

With the rapid development of the digital image editing tools, it is easy to tamper with the digital image even without leaving any perceptible traces. By means of duplicating regions to other places in the same image, copy-move forgery aims at enhancing the visual effect of the image or covering the truth \cite{TIFs2015JLi}. The goal of copy-move forgery detection is to determine the authenticity of the image by detecting the traces left by copy-move forgery. Copy-move forgery detection is one of the most actively investigated topics in image forensics \cite{TIFs2012VChristlein}.

In general, there are two main branches in copy-move forgery detection, one is block-based forgery detection, and the other is keypoint-based forgery detection \cite{TIFs2012VChristlein}. In the block-based copy-move forgery detection methods, overlapping image patches which contain raw or transformed pixels are extracted, and similar patches are sorted to seek traces of forgery \cite{TIFs2013SJRyu}. In the keypoint-based forgery detection methods, features of keypoints, e.g., the Scale-Invariant Feature Transform (SIFT) \cite{TIFs2011IAmerini} and the Speeded-Up Robust Features (SURF) \cite{JVCIR2015ESilva}, are adopted to represent the suspicious regions. In fact, the two kinds of methods both try to describe local features and evaluate the similarity of different patches. The major difference is that the block-based methods extract local features from abundant overlapping patches, while keypoint-based methods only consider patches of keypoints which are mostly located in high entropy regions. Although great progress has been made in recent researches, the adopted features for local patches mostly are hand-crafted features. Motivated by the good performance achieved by deep learning methods in computer vision tasks, we concentrate on the application of data-driven local descriptors in copy-move forgery detection. Convolutional Kernel Network (CKN) is a kind of data-driven patch-level descriptors which combines kernel methods and neural networks, achieving good performance and having excellent discriminative capability \cite{NIPs2014JMairal}. For the purpose of making use of data-driven features in copy-move forgery detection to improve its performance, we deliberate on the application and acceleration of CKN in this paper. In section \ref{sect:RelatedWork}, we will make a comprehensive analysis of the state-of-the-art copy-move forgery detection methods and data-driven descriptors.

In this paper, CKN is adopted to conduct copy-move forgery detection. In copy-move forgery detection, one of the important goals for feature representations is that the features should be invariant to particular transformations. In CKN, the invariance is encoded by a reproducing kernel which is demonstrated in the seminal work \cite{NIPs2014JMairal}. Different from conventional CNNs (Convolutional Neural Networks) which are learned either to represent data or for solving a classification task \cite{bayar2016deep}, CKN learns to approximate the kernel feature map on training data which is easy to train and robust to overfitting \cite{NIPs2014JMairal}. In \cite{YuanRao2016wifs}, Rao et al. proposed a method to conduct splicing detection and copy-move forgery detection using so-called local convolutional features, but they initialize the first layer of the network with the basic high-pass filter set and detect copy-move forgery in the same way as detecting splicing. Those so-called local features are designed to identify signature inconsistencies in various regions to locate forged regions, while the adopted data-driven convolutional local features in our work aim to find keypoints matches. CKN features can achieve competitive performance than conventional local features. Numerous experiments are conducted to demonstrate the effectiveness and robustness of the copy-move forgery detection based on CKN. Although promising results can be achieved by the proposed method, the original CKN is implemented on the CPU \cite{NIPs2016JMairal}, leading to low efficiency in forgery detection. In this paper, we reformulate CKN as a series of matrix computations and convolutional operations, which have enormous advantages to be implemented on GPU. Our GPU version of CKN can achieve high efficiency without significant compromising on effectiveness, making it possible to apply CKN to conducting copy-move forgery detection in batches. The contributions are two-fold:

\begin{list}{\textbullet}{%
\setlength{\topsep}{0pt} \setlength{\partopsep}{0pt}
\setlength{\parsep}{0pt} \setlength{\itemsep}{0pt}}
\item Firstly, a kind of data-driven convolutional local feature, i.e. CKN, is adopted to conduct keypoints matching in copy-move forgery detection. And CKN is reformulated and reimplemented on GPU to achieve high efficiency, making it possible to apply in copy-move forgery detection.
\item Secondly, appropriate preprocessing and postprocessing methods are adopted from \cite{TIFs2015JLi} to achieve copy-move forgery detection based on CKN. Although, the pipeline is the same as \cite{TIFs2015JLi}, two significant improvements are made to adjust to CKN: (1) a kind of keypoints distribution strategy based on oversegmentation is proposed to achieve homogeneous keypoints distribution; (2) a kind of adaptive oversegmentation method based on CNNs \cite{maninis2016convolutional}, i.e. COB (Convolutional Oriented Boundaries), is adopted to achieve better performance.
\end{list}

The rest of the paper is structured as follows: In Section \ref{sect:RelatedWork}, we discuss related work. In Section \ref{sect:Method}, we elaborate the proposed method. In Section \ref{sect:Experiments}, experiments are conducted to demonstrate the effectiveness and robustness of the proposed method. In Section \ref{sect:Conclusion}, we draw conclusions.

\section{Related Work}
\label{sect:RelatedWork}

The amount of literature relating to copy-move forgery detection or data-driven descriptors is immense, but the combination of both is rare. In this section, we will therefore discuss these two directions separately.

\begin{figure*}[htp]
\centering
  \centerline{\includegraphics[width=11cm]{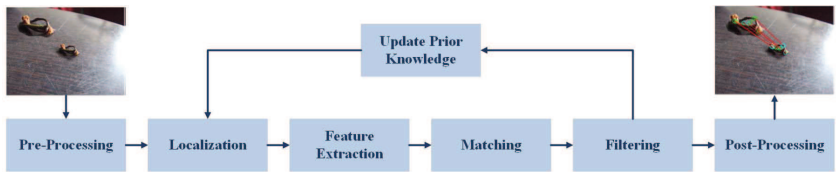}}
  \caption{Common framework of the copy-move forgery detection methods.}
  \label{fig:framework}
\end{figure*}

\textbf{Copy-move forgery detection:} During the last decades, various methods have been proposed to detect the copy-move forgery, and those methods can be broadly divided into two categories, namely block-based methods and keypoint-based methods. Referring to the workflow of common copy-move forgery detection methods concluded in \cite{TIFs2012VChristlein}, with two kinds of methods both considered, a common framework of copy-move forgery detection can be depicted as Fig. \ref{fig:framework}.

Firstly, the input image is preprocessed, the major preprocessing techniques include, e.g., the combination of color channels to generate a gray-scale image that can work properly for a given approach, or segmentation which aims at reducing the computational complexity and enhancing the detection accuracy, etc. In \cite{TIFs2015JLi}, the method firstly segments the image into patches by SLIC, and the keypoints matching is conducted under the restriction of the generated patches to detect copy-move regions. Similarly, in \cite{TIFs2015CMPun}, they proposed a kind of segmentation method called Adaptive Over-Segmentation algorithm to divide the host image into non-overlapping and irregular blocks adaptively. Similar to keypoint-based methods, feature points are extracted from each block to represent as the block feature.

Secondly, the localization means finding the patch centers, namely overlapping blocks of squared sizes or points of interest. Searching all possible locations and shapes can seldom miss the duplicated regions. However the computational complexity is almost unacceptable. Meanwhile, false matched areas are inevitable. On the other hand, conventional interest point detectors can only detect the points in high contrast regions, and may neglect smooth areas. In \cite{TIFs2016MZandi}, Zandi et al. proposed a novel interest point detector, which takes use of the advantages of both block-based and keypoint-based methods. By adopting the new detector, the low contrast regions can be detected.

As for the feature extraction procedure, we introduce it from two parts, keypoint-based algorithms and block-based algorithms. In keypoint-based methods, two kinds of feature extraction algorithms are commonly adopted, SIFT and SURF. Although a variety of preprocessing and postprocessing methods are used, the feature extraction procedures are almost the same. In another words, the existing keypoint-based methods differ mostly in the kind of interest points taken into consideration and in the matching policy used. In \cite{TIFs2010XPan,TIFs2011IAmerini,TIFs2015JLi,TIFs2015CMPun}, SIFT is chosen as the feature extraction method. While in \cite{shivakumar2011detection,JVCIR2015ESilva}, SURF is adopted. In \cite{TIFs2015EArdizzone}, the features of triangles are computed and compared instead of blocks or single points, and the triangles are built onto those keypoints, in which three keypoints detection methods are tested, namely, SIFT, SURF and Harris.

In block-based methods, many kinds of features have been adopted to describe the overlapping blocks, e.g.,  quantized DCT (Discrete Cosine Transform) coefficients adopted in \cite{fridrich2003detection}, PCA (Principal Component Analysis) in \cite{popescu2004exposing}, blur moment invariants with PCA for dimensional reduction in \cite{mahdian2007detection}, DWT (Discrete Wavelet Transform) and SVD (Singular Value Decomposition) \cite{li2007sorted}, Discrete Wavelet Transform or Kernel Principal Component Analysis \cite{bashar2010exploring}, Zernike moments \cite{TIFs2013SJRyu}, FMT (Fourier Mellin Transform) \cite{TIFs2015DCozzolino}, PCT (Polar Cosine Transform) \cite{li2013image}, LBP (Local Binary Patterns) \cite{li2013efficient}, etc. Although some features, e.g., DCT, PCA, SVD, etc., are mostly robust against JPEG compression, additive noise and blurring, and some features, e.g., FMT and LBP, are rotation invariant, those features are not simultaneously robust to scale, compression and rotation operations.

In the matching stage, similar patches will be detected. In the newly proposed method in \cite{TIFs2015DCozzolino}, PatchMatch algorithm is adopted to conduct feature matching with high efficiency, while Zernike Moments (ZM), Polar Cosine Transform (PCT) and Fourier-Mellin Transform (FMT) are considered for feature extraction. After the matching stage, it is inevitable that there are spurious pairs. The filtering stage is designed to remove those spurious pairs. In \cite{TIFs2016MZandi}, a novel filtering algorithm was proposed which can effectively prune the falsely matched regions, besides their newly proposed interest point detector as above mentioned.

In the post-processing stage, some simple processes, e.g., morphological operations, are mostly employed to only preserve matches that exhibit a common behavior. To refine the results further, some newly proposed methods update the matching information using the achieved knowledge from the previous iterations \cite{JVCIR2015ESilva,TIFs2015JLi}. In \cite{TIP2016AFerreira}, Ferreira et al. combined different properties of copy-move detection approaches, and modeled the problem on a multiscale behavior knowledge space, which encodes the output combinations of different techniques as the priori probabilities considering multiple scales of the training data.

Although various methods were proposed recently, leading to tremendous progress in copy-move forgery detection, few work has been conducted on the optimization of feature extraction. In the state-of-the-art methods, conventional hand-crafted descriptors (e.g., LBP, ZM, PCT, FMT, SIFT, SURF, etc.\cite{TIFs2012VChristlein,TIP2016AFerreira}) are widely adopted. Motivated by the great advance of deep learning methods in computer vision tasks \cite{NIPs2012AKrizhevsky,NIPs2015SRen,arXiv2016KHe}, we adopt a kind of data-driven local convolutional feature, namely CKN, to conduct copy-move forgery detection.

\textbf{Data-driven descriptors:} The algorithms adopted in copy-move forgery detection are mainly borrowed from the computer vision tasks, such as image classification \cite{CVPR2009JYang}, object detection \cite{PAMI2010PFFelzenszwalb} and image retrieval \cite{PAMI2000AWMSmeulders}, etc. Recent advances in computer vision tasks have been greatly promoted by the quick development of the GPU technologies and the success of convolutional neural networks (CNN) \cite{NIPs2012AKrizhevsky}. Different from conventional formulations of image classification based on local descriptors and VLAD \cite{PAMI2012HJegou} etc., the newly proposed image classification methods based on CNN adopt an end-to-end structure. Deep networks naturally integrate low/mid/high level features \cite{ECCV2014MDZeiler} and classifiers in an end-to-end multilayer fashion, and the levels of features can be enriched by the number of stacked layers (depth). Typical convolutional neural networks, e.g., AlexNet \cite{NIPs2012AKrizhevsky}, VGG \cite{ICLR2015KSimonyan}, ResNet \cite{arXiv2015KHe,arXiv2016KHe}, and ResNeXt \cite{arXiv2016SXie} etc., have greatly improved the performance on the tasks of image classification and object detection \cite{NIPs2015SRen}, etc. Features output by above-mentioned CNNs' intermediate layers can be regarded as image-level descriptors or so-called global features. Those global features are designed to reinforce inter class difference while neglect the intra class difference. And those deep learning based methods proposed for computer vision tasks can not directly be used in copy-move forgery detection which aims to find similar regions undergoing rotation, resizing or deformation.

Inspired by the expressive feature representations output by image-level CNNs, the question of whether suitable patch-level descriptors could be derived from such architectures has been raised \cite{ICCV2015MPaulin}, which aims at substituting data-driven descriptors for hand-crafted patch-level descriptors. Recently, several deep local descriptors were proposed and can achieve promising performance on patch matching and patch classification \cite{arXiv2014PFischer,arXiv2014ESimoSerra,NIPs2014JMairal,NIPs2016JMairal}. In \cite{arXiv2014PFischer}, a patch and patches generated by conducting different transformations on it are regarded as the same class, and the network is trained on those surrogate class labels. In \cite{arXiv2014ESimoSerra}, a siamese architecture of CNN is adopted and trained on matching/non-matching pairs. In \cite{NIPs2014JMairal}, the authors proposed a reproducing kernel which produces multi-layer image representation and trained the network without supervision which is called Convolutional Kernel Network (CKN). They bridge the gap between kernel methods and neural networks and aim at reaching the best of both worlds. In \cite{NIPs2016JMairal}, a supervised version of CKN was proposed and trained in an end-to-end manner. CKN has been tested on numerous datasets and also been adopted to conduct image retrieval in \cite{IJCV2017Paulin,ICCV2015MPaulin}, achieving competitive results with the state-of-the-arts. For its remarkable discriminative capability and invariance properties, we introduce CKN for copy-move forgery detection.

\section{Method}
\label{sect:Method}

\begin{figure*}[htp]
\centering
  \centerline{\includegraphics[width=12cm]{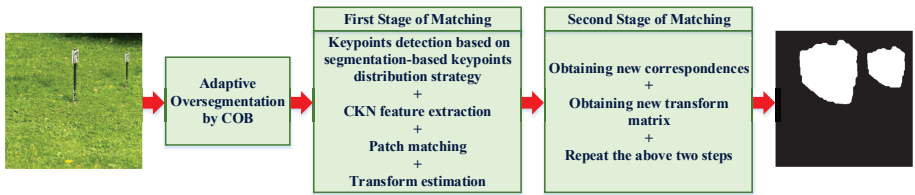}}
  \caption{The framework of the proposed copy-move forgery detection method based on CKN.}
  \label{fig:frameworkour}
\end{figure*}

The pipeline of the proposed copy-move forgery detection method originates from the work proposed by Li et al. \cite{TIFs2015JLi}, the framework of the proposed method is shown in Fig. \ref{fig:frameworkour}. As introduced in Section \ref{sect:introduction}, three significant changes are made: data-driven local descriptors (i.e. CKN), the segmentation-based keypoints distribution strategy and adaptive GPU-based oversegmentation (i.e. COB). Thus, this section is organized as follows: Firstly, we introduce the theoretic derivation of Convolutional Kernel Network; Secondly, we introduce the computation procedure of the CKN-grad and analysis its feasibility of implementing on GPU; Finally, we make an introduction on how to conduct forgery detection making use of CKN features, with the explanation of the proposed segmentation-based keypoints distribution strategy and the adopted adaptive GPU-based oversegmentation.

\subsection{Convolutional Kernel Network}
\label{sect:CKN}

\begin{figure*}[htp]
\centering
  \centerline{\includegraphics[width=11cm]{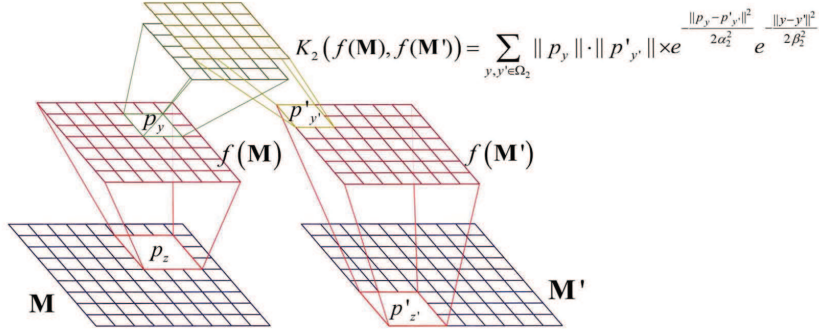}}
  \caption{The two-layer convolutional kernel architecture.}
  \label{fig:CKN}
\end{figure*}

Convolutional Kernel Network (CKN) is designed to output patch descriptors, and the input patches can be generated by keypoint detectors (e.g. DoG \cite{TIFs2011IAmerini}). Let $\mathbf{M}$ and $\mathbf{M}'$ denote two patches of size $m\times m$, and $\Omega = \{1,\cdots,m\}^2$ is the pixel locations set. $p_z$ denotes the sub-patch from $\mathbf{M}$ centered at location $z\in\Omega$ with a fixed sub-patch size (resp. $p'_{z'}$ denotes the sub-patch from $\mathbf{M}'$). In the implementation, the sub-patches near the border of $\mathbf{M}$ which have values outside of $\Omega$ are discarded without padding. The convolutional kernel between $\mathbf{M}$ and $\mathbf{M}'$ is defined as:
\begin {equation}\label{eq:singlelayerkernel}
K_1(\mathbf{M},\mathbf{M}')=\sum_{z,z'\in\Omega}{||p_z||||p'_{z'}||e^{-\frac{||z-z'||^2}{2{\beta_1}^2}}e^{-\frac{||\tilde{p}_z-\tilde{p}'_{z'}||^2}{2{\alpha_1}^2}}}
\end {equation}
where $\beta_1$ and $\alpha_1$ denote smoothing parameters of Gaussian kernels, $||\cdot||$ denotes $L_2$ norm, and $\tilde{p}_z:=(1/||p_z||)p_z$ which is the $L_2$-normalized version of the sub-patch $p_z$, and $\tilde{p}'_{z'}$ is the $L_2$ version of $p'_{z'}$. Thus, the feature representation of a patch is defined by the convolutional kernel. For that the kernel is a match kernel, a tunable level of invariance can be offered through the choice of hyperparameters, producing hierarchical convolutional representations \cite{NIPs2014JMairal}.

To compute formula (\ref{eq:singlelayerkernel}), the approximation procedure can be denoted as:
\begin {equation}\label{eq:kernel1approximation}
K_1(\mathbf{M},\mathbf{M}')\approx\sum_{u\in\Omega_{1}}{g_1(u;\mathbf{M})^Tg_1(u;\mathbf{M}')}
\end {equation}
\begin {equation}\label{eq:g1}
g_1(u;\mathbf{M}):=\sum_{z\in\Omega}e^{-\frac{||u-z||^2}{2{\beta_1}^2}}h_1(z;\mathbf{M}),u\in\Omega_{1}
\end {equation}
\begin {equation}\label{eq:h1}
h_1(z;\mathbf{M}):=||p_z||\lbrack\sqrt{\eta_j}e^{-\frac{||w_j-\tilde{p}_z||^2}{{\alpha_1}^2}}\rbrack_{j=1}^{n_1},z\in\Omega
\end {equation}
where $\Omega_{1}$ is the subset of $\Omega$, $w_j$ and $\eta_j$ are the learned parameters. There are two distinct approximations: 1) one is in the subsampling defined by $|\Omega_{1}|\leq|\Omega|$ that corresponds to the stride of the pooling operation in CNN; 2) the other is in the embedding of the Gaussian kernel of the subpatches:
\begin {equation}\label{eq:h1h1}
||p_z||||p'_{z'}||e^{-\frac{||\tilde{p}_z-\tilde{p}'_{z'}||^2}{2{\alpha_1}^2}}\approx h_1(z;\mathbf{M})^Th_1(z';\mathbf{M}')
\end {equation}

Since $K_1(\mathbf{M},\mathbf{M}')$ is the sum of the match-kernel terms, sampling $n$ pairs of sub-patches $\{(p_i,p'_i)\}_{i=1,\cdots,n}$, it can be approximated at sub-patch level by solving an optimization problem as follows:
\begin {equation}\label{eq:min}
min_{w_j,\eta_j}\sum_{i=1}^n\left(e^{-\frac{||\tilde{p}_i-\tilde{p}'_i||^2}{{\alpha_1}^2}}-\sum_{j=1}^{n_1}\eta_je^{-\frac{||w_j-\tilde{p}_i||^2}{{\alpha_1}^2}}e^{-\frac{||w_j-\tilde{p}'_i||^2}{{\alpha_1}^2}}\right)^2
\end {equation}

To get potentially better feature representation, the kernel can be overlaid by another kernel. In the single layer kernel, an approximation spatial map $f_1(\mathbf{M})$ is computed, where $\mathbf{M}$ denotes an input patch. Thus, a kernel $K_2$ can be defined in the same way as $K_1$. The two-layer convolutional kernel architecture is shown in Fig. \ref{fig:CKN}. For the ingenious design of CKN, it is difficult to draw out the theoretical roots in such a short piece, readers can kindly refer to \cite{NIPs2014JMairal,NIPs2016JMairal} for the introduction of training procedures and detailed proofs. Of course, it is easy to understand the full procedure of forgery detection based on CKN, keeping in mind that CKN generates feature descriptors for patches.

\subsection{CKN-grad Computation}
\label{sect:CKNgrad}

\begin{figure*}[htp]
\centering
  \centerline{\includegraphics[width=11cm]{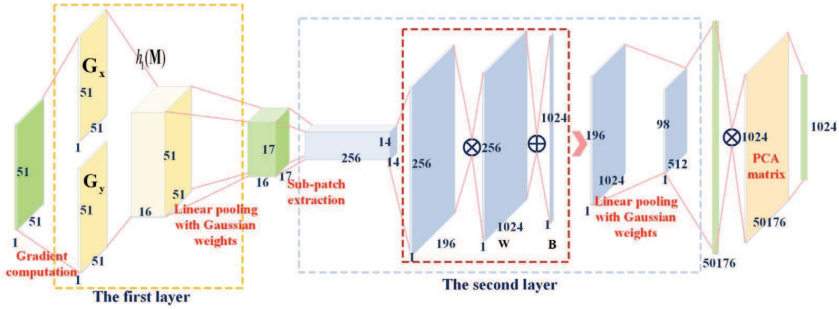}}
  \caption{The computation procedure of CKN-grad.}
  \label{fig:CKNGRAD}
\end{figure*}

In this paper, a two-layer structure CKN called \textbf{CKN-grad} is adopted, in which the input is the gradient along each spatial dimension (the input multi-channel image is transformed to one channel to compute the gradient). The size of the input patch is $51\times51$ (which means $m=51$ in $\Omega$), and the size of the sub-patch is set as $1\times1$. The input at location $z$ is $p_z=(G_{z_x},G_{z_y})$, where $G_{z_x}$ and $G_{z_y}$ are the gradients along axis $x$ and axis $y$ respectively. Because the input features are normalized, the inner part of the match kernel $||\tilde{p}_z-\tilde{p}'_{z'}||$ is directly linked to the cosine of the angle between the two gradients, see \cite{NIPs2014JMairal}. So the approximation of kernel $K_1$ is computed as:
\begin {equation}\label{eq:apl1}
e^{-\frac{||\tilde{p}_z-\tilde{p}'_{z'}||^2}{2{\alpha_1}^2}}\approx \sum_{j=1}^{n_1}{\varphi_1(j;p_z)\varphi_1(j;p'_{z'})}
\end {equation}
\begin {equation}\label{eq:apl1}
\varphi_1(j;p_z)=\mathrm{exp}\left(-\frac{(\mathrm{cos}\theta_j-G_{z_x}/\rho)^2+(\mathrm{sin}\theta_j-G_{z_y}/\rho)^2}{{\alpha_1}^2}\right)
\end {equation}
where
\begin {equation}\label{eq:rho}
\rho=\sqrt{G_{z_x}^2+G_{z_y}^2}
\end {equation}
\begin {equation}\label{eq:alpha1}
\alpha_1=\sqrt{\left(1-\mathrm{cos}\left(\frac{2\pi}{n_1}\right)\right)^2+\mathrm{sin}\left(\frac{2\pi}{n_1}\right)^2}
\end {equation}
 and $\theta_j=2j\pi/n_1$, $j\in\{1,\cdots,n_1\}$ (we set $n_1=16$). Thus, given the input map $\mathbf{M}$, the input map before the convolution of Gaussian weights and pooling can be computed as:
 \begin {equation}\label{eq:h1M}
h_1(\mathbf{M})=\left(\left(\rho\cdot\varphi_1(j;p_z)\right)_{j=1}^{n_1}\right)_{z\in\Omega}
\end {equation}
 the output map $f(\mathbf{M})$ of the first layer is computed as:
\begin {equation}\label{eq:fM}
f(\mathbf{M})=\left(\mathrm{conv}\left(\mathbf{K}_g(\gamma_1),h_1(\mathbf{M})\right)\right)_{\hat{z}\in\Omega_{1}}
\end {equation}
where $\mathrm{conv}(\cdot)$ denotes the convolutional operation, $\mathbf{K}_g(\gamma_1)$ denotes the Gaussian kernel with a factor $\gamma_1$ (we set $\gamma_1=3$), $\Omega_{1}$ is obtained by subsampling $\Omega$ with the stride of $\gamma_1$. With the factor $\gamma_1$, $L_{k_1}=2\times\gamma_1+1$, the size of $\mathbf{K}_g(\gamma_1)$ is $L_{k_1}\times L_{k_1}$, and $\mathbf{K}_g(\gamma_1)$ is computed as:
\begin {equation}\label{eq:Kg}
\mathbf{K}_g(\gamma_1)=\left(\frac{\mathrm{k}_g(k_1,k_2,\gamma_1)}{\sum_{k_1}\sum_{k_2}\mathrm{k}_g(k_1,k_2,\gamma_1)}\right)_{L_{k_1}\times L_{k_1}}
\end {equation}
where $\mathrm{k}_g(k_1,k_2,\gamma_1)=\mathrm{exp}(-({k_1}^2+{k_2}^2)/2(\gamma_1/\sqrt{2})^2)$, $k_1$, $k_2$ are the relative coordinates to the center of the kernel $\mathbf{K}_g(\gamma_1)$. Thus, the output map of the first layer is a tensor of the size of $(m/\gamma_1)\times(m/\gamma_1)\times n_1=17\times17\times16$.

In the second layer, the input map is $f(\mathbf{M})$, the size of the sub-patch $p_y$ is $m_{p_y}\times m_{p_y} = 4\times4$, subsampling factor is set as $\gamma_2=2$ and $n_2=1024$. Omitting the borders, the number of input sub-patches is $(m/\gamma_1-m_{p_y}+1)\times (m/\gamma_1-m_{p_y}+1) \times n_1 = 14\times14\times16$. So the input of the second layer can be transformed to a matrix $\mathbf{M}_2$ of size $256\times196$. By conducting the approximation procedure introduced in \ref{sect:CKN}, the parameters $\mathbf{W}_i=\left(w_j\right)_{j=1}^{n_2}$ and $\boldsymbol{\eta}_i=\left(\eta_j\right)_{j=1}^{n_2}$ are learned, $i=1,\cdots,N$ ($N=256$). Trying to formulate it as the matrix computation, the parameters are transformed to a weight matrix $\mathbf{W}_{1024\times256}$ and bias $\mathbf{B}_{1024}$, the elements are computed as:
\begin {equation}\label{eq:W}
w_j=\frac{2w_j}{{\alpha_2}^2}
\end {equation}
\begin {equation}\label{eq:B}
b_j=\frac{log(\eta_j)}{2}-\frac{1+||w_j||^2}{{\alpha_2}^2}
\end {equation}
To reduce parameters, we set $b_j=\sum_{i=1}^{N}{b_i}/N$. Thus, we can get a 1024-dimensional vector $\mathbf{B}_{1024}$. $\mathbf{B}_{1024\times196}$ is a matrix with each column equal to $\mathbf{B}_{1024}$. Finally, the output map is computed as:
\begin {equation}\label{eq:B}
h_2(\mathbf{M}_2) = \mathbf{W}_{1024\times256} \times \mathbf{M}_2 + \mathbf{B}_{1024\times196}
\end {equation}
So the size of the output map is $1024\times196$. With the subsampling factor set as $\gamma_2=2$, the size of the final output map is $512\times98$ after linear pooling with Gaussian weights. The computation procedure of linear pooling with Gaussian weights is the same as formula (\ref{eq:fM}). Thus, the total dimension of the feature vector extracted by CKN-grad is $50176$. Then PCA (Principle component analysis) is adopted for dimensionality reduction. The hyperparameters and the PCA matrix both are obtained by training on the RomePatches dataset \cite{ICCV2015MPaulin}. Finally, a $1024$-dimensional feature vector can be extracted by CKN-grad from a patch of size $51\times51$.

\begin{figure}[htp]
  \centering
  \centerline{\includegraphics[width=6cm]{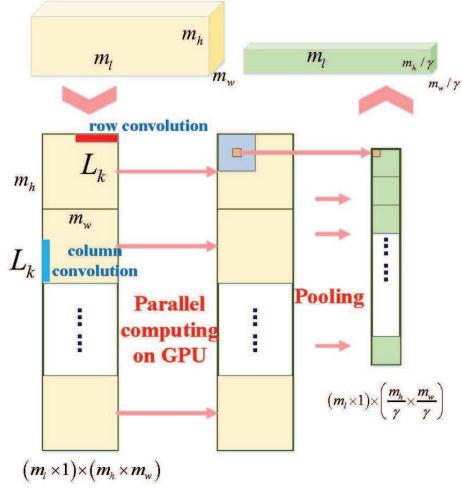}}
  \caption{GPU acceleration of linear pooling with Gaussian weights.}
  \label{fig:F6}
\end{figure}

As shown in Fig. \ref{fig:CKNGRAD}, the diagram of the CKN-grad structure and computation procedure is given. It can be clearly seen that the computation of CKN-grad is the process of numerous matrix computations in fact. Thus, it can be implemented on GPU directly which will be more efficient. Besides, in the procedure of linear pooling with Gaussian weights, a kind of acceleration method is adopted to conduct convolution, as shown in Fig. \ref{fig:F6}. In the first layer, a tensor of size $51\times51\times16$ is input to conduct linear pooling with Gaussian weights. There are two steps in this procedure: 1) convolutional computation with Gaussian weights, 2) pooling. In the original codes provided by the seminal work \cite{NIPs2014JMairal}, $16$ matrixes of $51\times51$ are convolved separately along $16$ loops which are time consuming. We propose to use an acceleration procedure without loop. With the subsampling factor set as $\gamma_1=3$, the tensor needs to be convolved with a kernel of $L_{k_1}\times L_{k_1}=7\times7$. Conventionally, it is equivalent to the combination of row convolution and column convolution. To reduce the communication between the CPU and GPU, the tensor is transformed to a matrix of size $(16\times1)\times(51\times51)$ (see Fig. \ref{fig:F6}) which is transferred into GPU altogether. On GPU, each block of size $51\times51$ is convolved separately and parallelly by the row vector and column vector. Then, a pooling operation is conducted parallelly on the map after row convolution and column convolution. In the second layer, the input matrix is firstly transformed to a tensor of size $14\times14\times1024$. Then the same computation is conducted. As shown in Fig. \ref{fig:F6}, it is the computation process of the linear pooling with Gaussian weights on GPU.

\subsection{Forgery Detection}
\label{sect:CMFD}

As shown in Fig. \ref{fig:frameworkour}, the suspicious image firstly is segmented into an abundant number of regions, i.e. oversegmentation. In the original work proposed by Li et al. \cite{TIFs2015JLi}, the SLIC (Simple Linear Iterative Clustering) algorithm \cite{achanta2012slic} is adopted to conduct oversegmentation, and SLIC is a popular superpixel segmentation method which has been widely used in saliency detection \cite{borji2014salient}, semantic segmentation \cite{mostajabi2015feedforward} and many other computer vision tasks. The SLIC algorithm adapts a k-means clustering approach to efficiently generate the superpixels, and it adheres to the boundaries very well. However, the initial size of the superpixels in SLIC is decided empirically, and of course difficult to decide \cite{liu2015saliency}. In another work of copy-move forgery detection \cite{TIFs2015CMPun}, they also employed the SLIC segmentation method for image blocking, and proposed a method to determine the initial size of the superpixels adaptively based on the texture of the host image.

\begin{figure*}[htp]
\centering
  \centerline{\includegraphics[width=13cm]{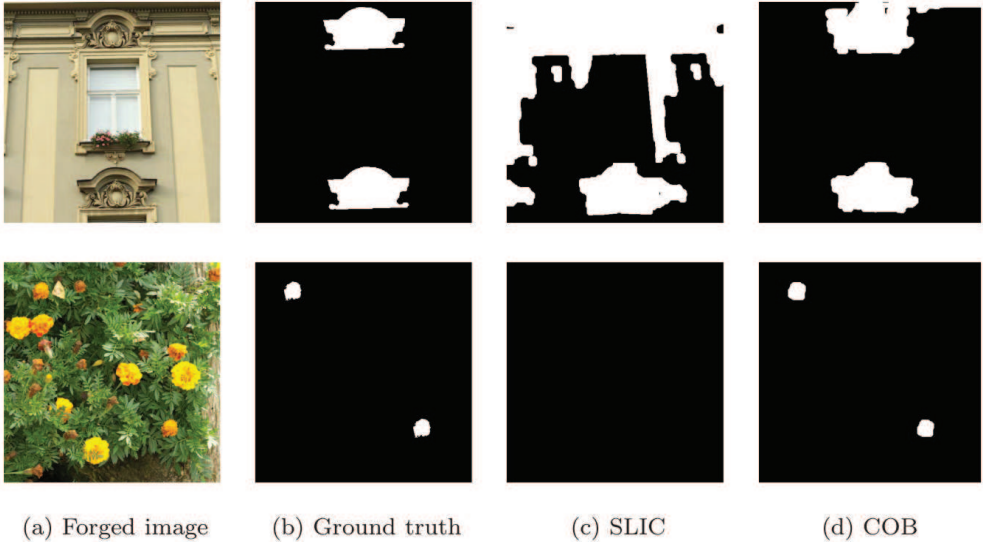}}
  \caption{The comparison between copy-move forgery detection based on SLIC and COB.}
  \label{fig:visualshown}
\end{figure*}

In this paper, we adopt another kind of segmentation method, namely, COB (Convolutional Oriented Boundaries) \cite{maninis2016convolutional} which can produce multiscale oriented contours and region hierarchies. COB requires a single CNN forward pass for contour detection and uses a novel sparse boundary representation for hierarchical segmentation, giving a significant leap in efficiency and performance. In fact, COB and its precursor, i.e. MCG \cite{pont2017multiscale} all originate from gPb-owt-ucm \cite{arbelaez2011contour}. Though gPb-owt-ucm can achieve excellent performance on accuracy, it is excluded in the original work of Li et al. \cite{TIFs2015JLi} for its complexity and inefficiency. In \cite{TIFs2015JLi}, they adopt SLIC, and set the region size empirically as Table \uppercase\expandafter{\romannumeral2} in \cite{TIFs2015JLi}. With the help of recent developed CNN features and HED edge detection\cite{xie2015holistically}, COB is much more efficient (demonstrated as Table 2 in \cite{maninis2016convolutional}) and accurate (demonstrated as Fig. 8 in \cite{maninis2016convolutional}) than gPb-owt-ucm, MCG and many other segmentation methods. COB can generate superpixels automatically and adaptively based on the detected edges. So, in the first step of oversegmentation in our method, we make use of COB. As shown in Fig. \ref{fig:visualshown}, we simply replace the segmentation method in \cite{TIFs2015JLi}, and it can be clearly seen that accurate and adaptive segmentation is even helpful to achieve more accurate detection or avoid missing detection. Besides, COB can also be used as a kind of object proposal method which can generate a set of segmentations which may contain entire objects. As discussed in \cite{TIFs2015JLi}, the image should be segmented into small patches, each of which is semantically independent to the others. The multiscale oriented contours and region hierarchies of COB may be helpful to solve this problem, and we leave it for future work.

After image segmentation, the first stage of matching is conducted, and the first step is keypoints detection. In the original work \cite{TIFs2015JLi}, they employed DoG (difference of Gaussian) with a fixed threshold ($0.004$ in their implementation). In \cite{yang2017copy}, Yang et al. defined the keypoints uniformity measurement (KUM) value, which denotes the keypoints
distribution level. With iterations of keypoints detection, they get an appropriate threshold for DoG which the KUM value of the detected keypoints is lower than a fixed value ($0.3$ in \cite{yang2017copy}). In this paper, motivated by the homogeneous distribution of superpixels, we propose a kind of segmentation-based keypoints distribution strategy. In our method, DoG keypoint detection is firstly conducted with a very low threshold (we set the lowest value $0$). And let $\mathbf{L}$ denotes the label matrix generated by COB, $L_i$ denotes the superpixel $i$, $n$ denotes the number of superpixels, $\mathbf{K}$ denotes the detected keypoints, $m$ denotes the number of keypoints, and $\mathbf{S}$ denotes the corresponding scores of keypoints which can be generated by DoG. Thus, the pseudocode of the proposed segmentation-based keypoints distribution strategy can be formulated as Algorithm \ref{alg:sbkds}. As shown in Algorithm \ref{alg:sbkds}, according the size proportion of each superpixel, we compute the standard number of keypoints of each superpixel. If the number of detected keypoints is less than the standard number in the superpixel, we output the detected keypoints in this superpixel directly. Otherwise, we sort the keypoints according to their scores, and output a standard number of keypoints. Meanwhile, to avoid missing those "good" keypoints, we also output the keypoints which have larger scores than a threshold score (we set as the median of scores, i.e., $\lambda=0.5$).

\begin{algorithm}[htp]
\caption{Segmentation-based keypoints distribution strategy}
\label{alg:sbkds}
\begin{algorithmic}[1]
\REQUIRE $\mathbf{L}=\{L_i|i=1,2,\cdots n\}$, $\mathbf{K}=\{k_j|j=1,2,\cdots m\}$ and $\mathbf{S}=\{s_j|j=1,2,\cdots m\}$ \\
\STATE $threshold\_score=s_{round(\lambda \times m)}$
\STATE $\mathbf{K}_o=\{\ \}$
\FOR{$i=1$ to $n$}
\STATE $standard\_keypoint\_num_{i}= m \times size(L_i)/ size(\mathbf{L})$
\STATE $\mathbf{K}_{i}= \{k_t\ in\ L_i |t=1,2,\cdots m_i\}$
\STATE $\mathbf{S}_{i}= \{s_t\ corresponding\ to\ k_t |t=1,2,\cdots m_i\}$
\IF {$m_i<=standard\_keypoint\_num_{i}$}
\STATE $\mathbf{K}_o = \mathbf{K}_o \cup \mathbf{K}_{i}$
\ELSE
\STATE Sort $\mathbf{K}_{i}$ according to $\mathbf{S}_{i}$:\\
 $\mathbf{K}^s_{i}= \{k^s_t|t=1,2,\cdots m_i\}$ \\
 $\mathbf{S}^s_{i}= \{s^s_t\ corresponding\ to\ k^s_t |t=1,2,\cdots m_i\}$
\STATE $\mathbf{K}_o = \mathbf{K}_o \cup \{k^s_t|t=1,\cdots standard\_keypoint\_num_{i}\}$
\FOR{$t=standard\_keypoint\_num_{i}+1$ to $m_i$}
\IF {$s^s_t>threshold\_score$}
\STATE $\mathbf{K}_o = \mathbf{K}_o \cup \{k^s_t\}$
\ENDIF
\ENDFOR
\ENDIF
\ENDFOR
\ENSURE $\mathbf{K}_o$
\end{algorithmic}
\end{algorithm}

Once the keypoints $\mathbf{K}_o$ are detected, the CKN features can be extracted from those detected keypoints as introduced in Section \ref{sect:CKNgrad}. Then the rest steps are the same as the original work proposed in \cite{TIFs2015JLi}, which can be concluded as following steps: (1) the detection of the suspicious pairs of regions which contain many similar keypoints. Specifically, in each region, for each keypoint, we search its $K$ nearest neighbors that are located in the other regions, with constructing a k-d tree to decrease the complexity of searching $K$ nearest neighbors. (2) After the step (1), suspicious pairs of regions are detected, then we estimate the relationship between these two regions in terms of a transform matrix by conducting RANSAC method. (3) For the reason of that a limited number of keypoints cannot resist the possible errors in keypoint extraction, a so-called second stage of matching is conducted to eliminate false alarm regions. As shown in Fig. \ref{fig:frameworkour}, this stage consists of two steps, namely obtaining new correspondences and obtaining new transform matrix. The estimation of the transform matrix is refined via an EM-based algorithm. Due to space limitations, we can not explain the theoretical derivation of these steps in detail, readers can kindly refer to the original work \cite{TIFs2015JLi} for help.

\section{Experiments}
\label{sect:Experiments}

In this paper, the contributions are two-fold: the data-driven convolutional local descriptor adoption, and the appropriate formulation of CKN-based copy-move forgery detection to achieve the state-of-the-art performance, which have been discussed in Section \ref{sect:introduction}. Thus, we conduct the experiments from two aspects: CKN evaluation (Section \ref{sect:CKNEVAL}) and comparison with other methods (Section \ref{sect:COM}). In Section \ref{sect:CKNEVAL}, we try to demonstrate the effectiveness and robustness of our GPU-based CKN; In Section \ref{sect:COM}, we try to demonstrate that the proposed method can achieve the state-of-the-art performance, and it is robust to different kinds of attacks.

\subsection{CKN Evaluation}
\label{sect:CKNEVAL}

As introduced in Section \ref{sect:CKNgrad}, the CKN is reformulated as a series of matrix computations and convolutional operations. To demonstrate the effectiveness and robustness of our GPU version of CKN, experiments are conducted from two parts: (1) the comparison between CKN and the GPU version of CKN by conducting patch retrieval to demonstrate its efficiency; (2) the performance of CKN in the field of copy-move forgery detection. In this part, we adopt the conventional framework of copy-move forgery detection based on SIFT \cite{TIFs2011IAmerini} for fair comparison. In another word, we simply replace the feature extraction method in \cite{TIFs2011IAmerini} to make a fair comparison between SIFT and CKN. In this formulation, suspicious areas are detected based on feature matching and hierarchical clustering without further preprocessing or postprocessing. Thus, the effectiveness and robustness of extracted features can be demonstrated.

As discussed in Section \ref{sect:CKN}, the goal of the CKN optimization is to approximate the kernel feature map on the training data, and the training process is totally the same as the original work of CKN for image retrieval \cite{ICCV2015MPaulin}, in which the network is trained on the train split of RomePatches based on the SGD optimization. Readers can kindly refer to \cite{ICCV2015MPaulin} for the detailed introduction of training procedures. The training process of CKN is not the concern of this work, and the contribution for CKN in this work is the GPU-based formulation and acceleration, as introduced in Section \ref{sect:CKNgrad}. The training sets of RomePatches are totally different from the test images for copy-move forgery detection evaluation. In another word, the parameters of CKN learned from RomePatches can be applied to different conditions which will be demonstrated in the following experiments.

The CKN codes provided by the seminal work \cite{NIPs2016JMairal} are implemented based on CPU , which results in low efficiency, as shown in Table \ref{table:patchretrieval}. The low efficiency seriously prevents the application of CKN. Especially, in the domain of copy-move forgery detection, thousands of patches are detected in a single image, which need further feature extraction. Thus, we reformulate CKN making it possible to implement based on GPU and accelerate the feature extraction. To demonstrate the effectiveness and efficiency of our GPU version of CKN, experiments are conducted on the RomePatches dataset \cite{ICCV2015MPaulin}. The train set and the test set of RomePatches dataset both contain $10000$ patches, and the total feature extraction time of \emph{CKN-grad} (CPU version) and \emph{CKN-grad-GPU} is recorded. For comprehensive comparisons, we conduct the experiments on two machines: Machine (1) with Intel(R) Core(TM) i7-5930K CPU $@$ 3.50GHz, $64$GB RAM and a single GPU (TITAN X); Machine (2) with Intel(R) Xeon(R) CPU E5-2640 v2 $@$ 2.00GHz, $64$GB RAM and a single GPU (Tesla M40). Clearly, our GPU version of CKN, i.e. \emph{CKN-grad-GPU}, is more efficient. The speed of \emph{CKN-grad-GPU} is at least $8$ times of that of \emph{CKN-grad} on CPU ($16$ times on machine (2)), with only a little compromising on the accuracy of patch retrieval, which will be shown that there is no difference between \emph{CKN-grad} and \emph{CKN-grad-GPU} to conduct copy-move forgery detection in the next experiments.

\begin{table*}[!t]
\renewcommand{\arraystretch}{1.3}
\caption{Results for patch retrieval}
\label{table:patchretrieval}
\centering
\scriptsize
\begin{tabular}{c c c c c c }
\hline
\multirow{3}{*}{Algorithm} & \multirow{3}{*}{Machine} & \multicolumn{2}{c}{RomePatches train} &   \multicolumn{2}{c}{RomePatches test}   \\
 &  & \tabincell{c}{running\\ time(s)} & accuracy(\%) & \tabincell{c}{running\\ time(s)} & accuracy(\%) \\
\hline
CKN-grad & \multirow{2}{*}{(1)} & 99.74 & 92.06 & 101.73 & 86.98 \\
CKN-grad-GPU &  & 12.79 & 91.51 & 12.77 & 86.42 \\
\hline
CKN-grad & \multirow{2}{*}{(2)} & 244.47 & 92.06 & 244.98 & 86.98 \\
CKN-grad-GPU &  & 15.22 & 91.51 & 15.31 & 86.42 \\
\hline
\end{tabular}
\end{table*}

\begin{table*}[!t]
\renewcommand{\arraystretch}{1.3}
\caption{Copy-move forgery detection results on MICC-F220 dataset}
\label{table:F220}
\centering
\scriptsize
\begin{tabular}{c c | c c c c | c c}
\hline
Algorithm & metric & TN & TP & FN & FP & TPR & FPR \\
\hline
SIFT-origin & ward & 100 & 108 & 2 & 10 & 98.18\% & 9.09\%  \\
SIFT-origin & single & 104 & 99 & 11 & 6 & 90.00\% & 5.45\%  \\
SIFT-origin & centroid & 101 & 108 & 2 & 9 & 98.18\% & 8.18\%  \\
\hline
SIFT-VLFeat & ward & 97 & 110 & 0 & 13 & 100.00\% & 11.82\%  \\
SIFT-VLFeat & single & 104 & 101 & 9 & 6 & 91.82\% & 5.45\%  \\
SIFT-VLFeat & centroid & 98 & 110 & 0 & 12 & 100.00\% & 10.91\%  \\
\hline
CKN-grad & ward & 101 & 110 & 0 & 9 & 100.00\% & 8.18\% \\
CKN-grad & single & 103 & 96 & 14 & 7 & 87.27\% & 6.36\%  \\
CKN-grad & centroid & 102 & 109 & 1 & 8 & 99.09\% & 7.27\%  \\
\hline
CKN-grad-GPU & ward & 101 & 110 & 0 & 9 & 100.00\% & 8.18\% \\
CKN-grad-GPU & single & 103 & 96 & 14 & 7 & 87.27\% & 6.36\% \\
CKN-grad-GPU & centroid & 102 & 109 & 1 & 8 & 99.09\% & 7.27\% \\
\hline
\end{tabular}
\end{table*}

In order to demonstrate the effectiveness and robustness of CKN and the GPU version of CKN, elaborate experiments are conducted on two publicly available datasets: MICC-F220 and MICC-F2000. There are $220$ images in MICC-F220, in which $110$ images are original and $110$ images are tampered. In MICC-F2000, there are $2000$ images with $1300$ original images and $700$ tampered images. The detection performance is measured by the true positive rate (TPR) and false positive rate (FPR). Specifically, TPR$=$TP$/($TP$+$FN$)$ and FPR$=$FP$/($FP$+$TN$)$, where TP denotes true positive which means images detected as forged being forged, FN denotes false negative which means images detected as original being forged, FP denotes false positive which is the number of images detected as forged being original, and TN denotes true negative which is the number of original images detected as original.

First of all, experiments are conducted with $T_h$ and min cluster pts (which denotes the least number of pairs of matched points linking a cluster to another one) fixed. Ward, single, and centroid are three kinds of linkage metrics used to stop cluster grouping with the threshold ($T_h$), readers can refer to \cite{TIFs2011IAmerini} for details. Four kinds of methods are tested, namely, \emph{SIFT-origin}, \emph{SIFT-VLFeat}, \emph{CKN-grad} and \emph{CKN-grad-GPU}. The results of \emph{SIFT-origin} are generated by the codes provided by the seminal work \cite{TIFs2011IAmerini}, the patch size of \emph{SIFT-origin} is $16\times 16$. For fair comparison, in \emph{SIFT-VLFeat}, we extract SIFT features from patches of size $51\times 51$. In \emph{SIFT-VLFeat}, \emph{CKN-grad} and \emph{CKN-grad-GPU}, we adopt the codes provided by VLFeat to conduct patch extraction.

It can be seen from Table \ref{table:F220} that \emph{CKN-grad} and \emph{CKN-grad-GPU} can generate the same results without any difference, which means that there is no significant compromising of effectiveness by reformulating CKN on GPU, while the GPU version is more efficient. For this reason, a comprehensive comparison is conducted in the next part among \emph{SIFT-origin}, \emph{SIFT-VLFeat} and \emph{CKN-grad-GPU} (we consider that \emph{CKN-grad} and \emph{CKN-grad-GPU} are the same). In general, \emph{SIFT-VLFeat} and \emph{CKN-grad-GPU} can constantly achieve better performance than \emph{SIFT-origin}. For the reason of that the performance of SIFT can be influenced by the patch size \cite{simo2015discriminative}. Besides, the number of patches generated by VLFeat is not the same as the original codes \cite{TIFs2011IAmerini}, though they both adopt DoG algorithms. \emph{SIFT-VLFeat} and \emph{CKN-grad-GPU} can achieve similar TPRs, while \emph{CKN-grad-GPU} can achieve lower FPRs.

\begin{figure*}[htp]
\begin{minipage}[b]{0.3\linewidth}
  \centering
  \centerline{\includegraphics[width=4.5cm]{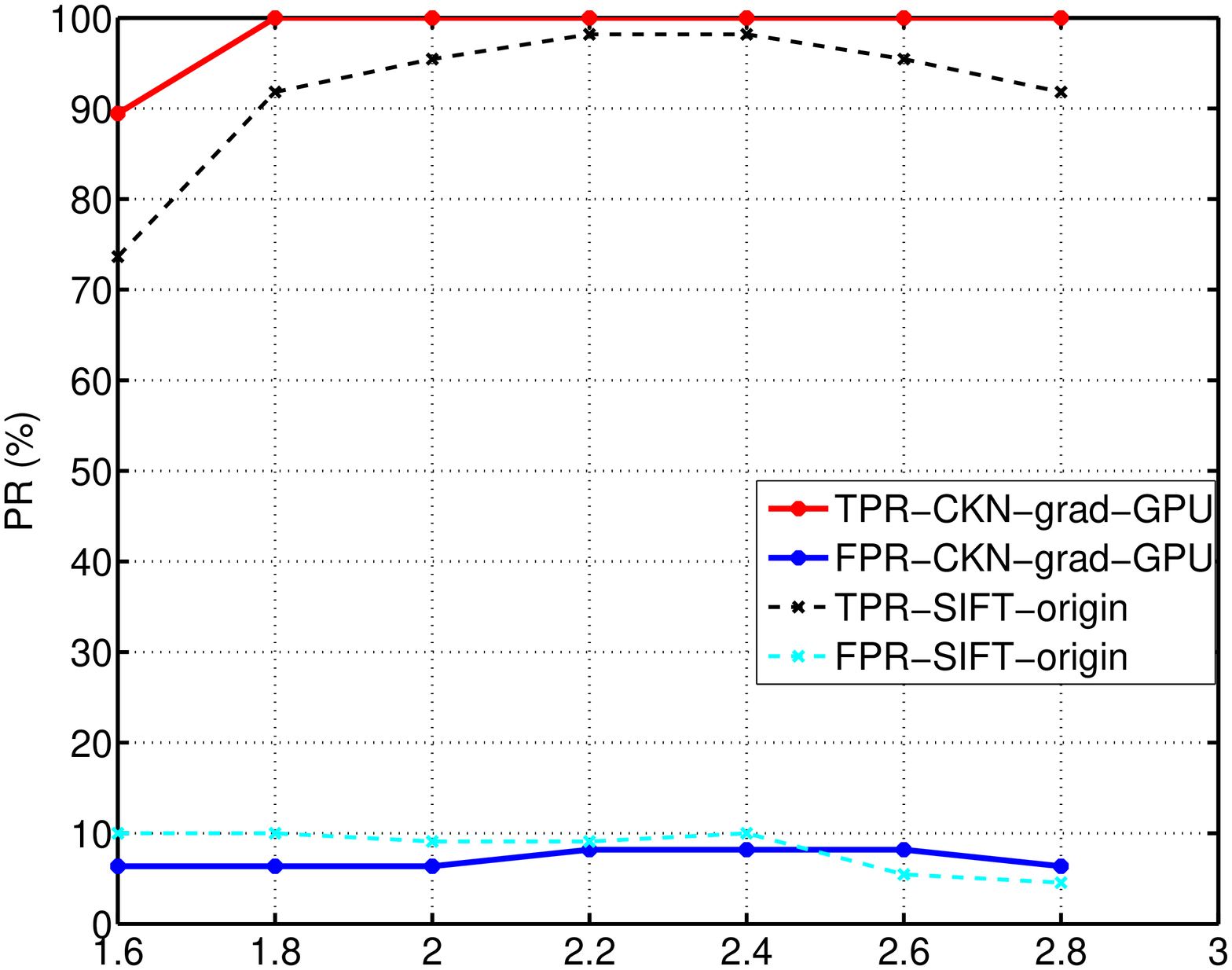}}
  \centerline{ }
\end{minipage}
\hfill
\begin{minipage}[b]{0.3\linewidth}
  \centering
  \centerline{\includegraphics[width=4.5cm]{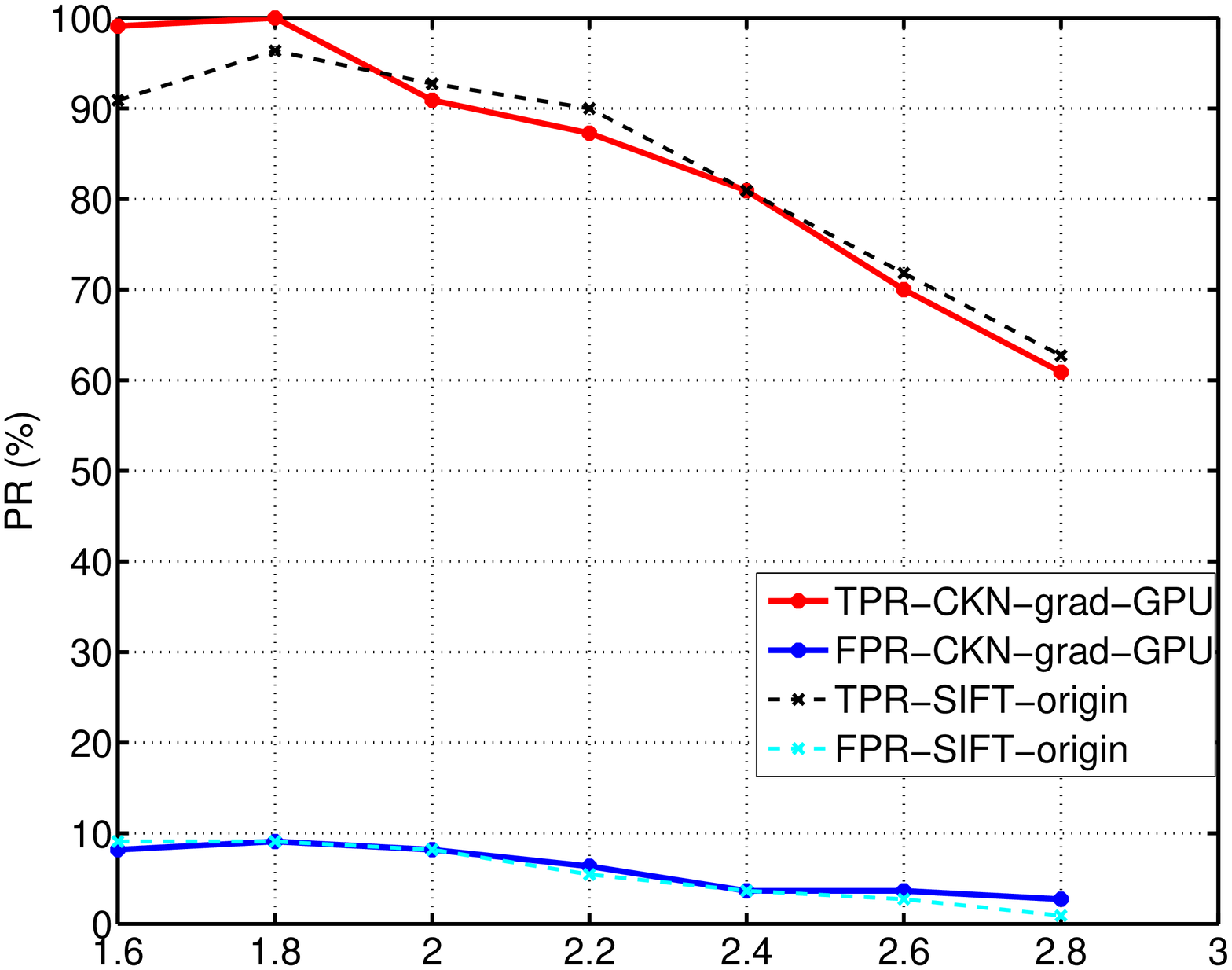}}
  \centerline{ }
\end{minipage}
\hfill
\begin{minipage}[b]{0.3\linewidth}
  \centering
  \centerline{\includegraphics[width=4.5cm]{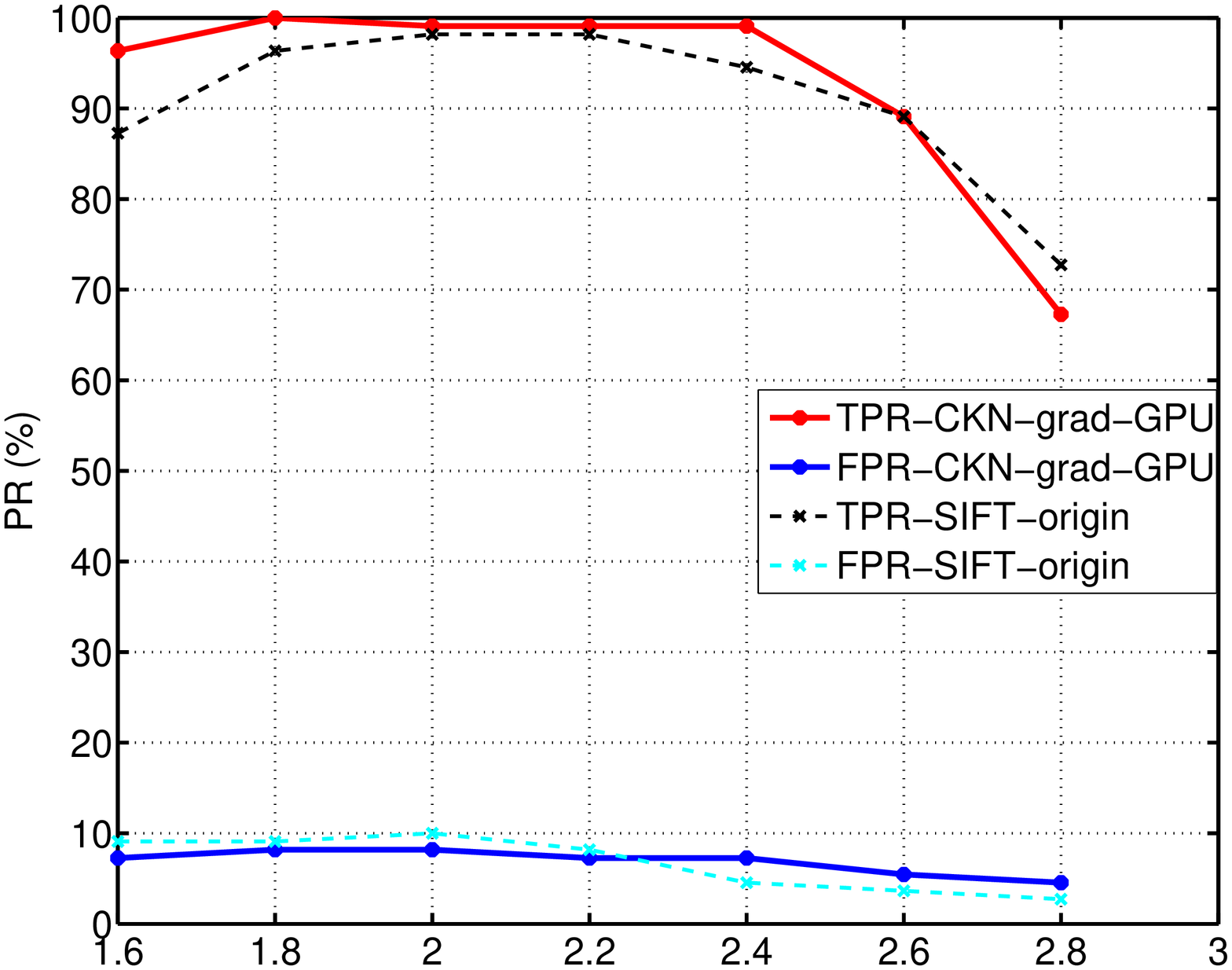}}
  \centerline{ }
\end{minipage}
\vfill
\begin{minipage}[b]{0.3\linewidth}
  \centering
  \centerline{\includegraphics[width=4.5cm]{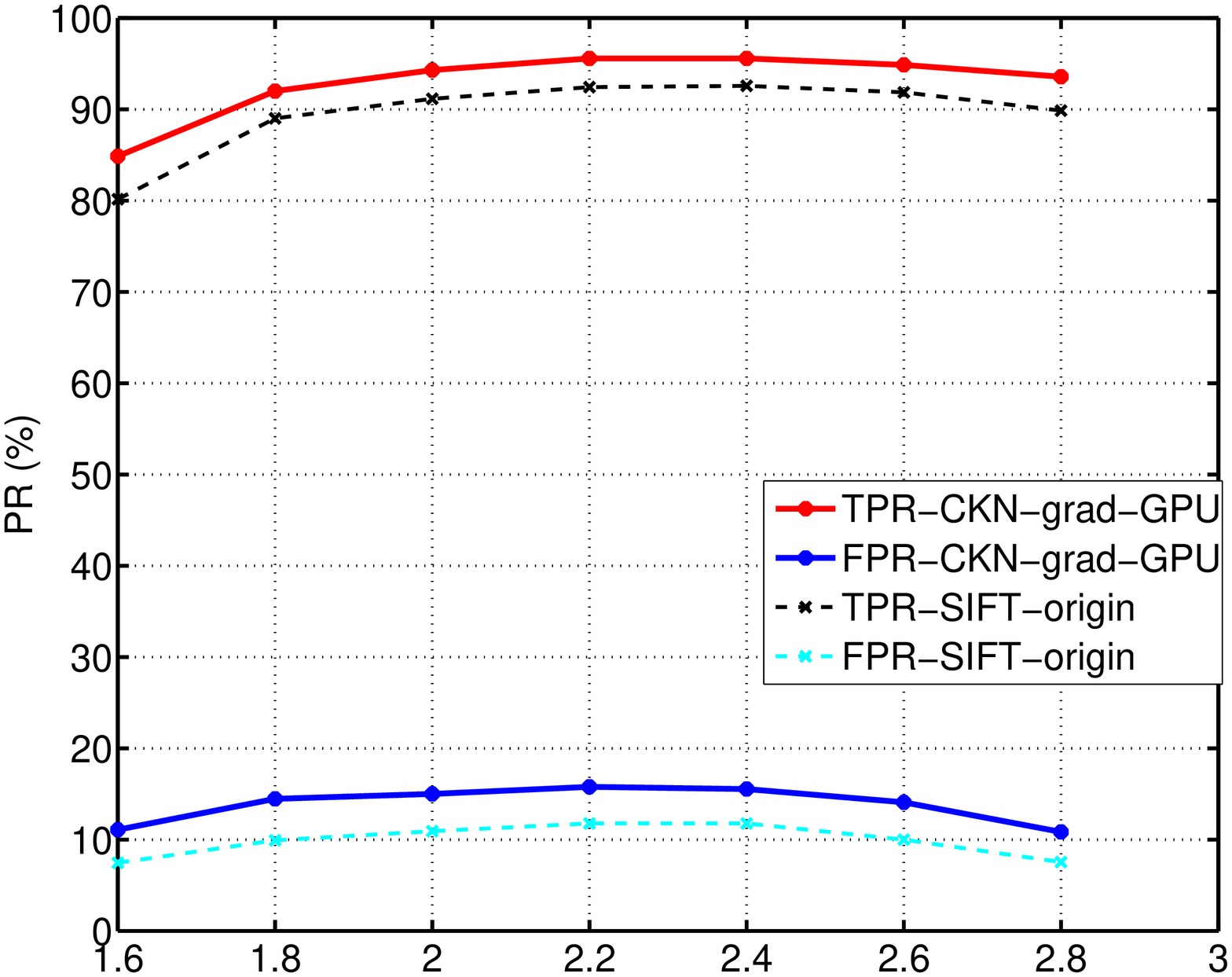}}
  \centerline{(a)ward}
\end{minipage}
\hfill
\begin{minipage}[b]{0.3\linewidth}
  \centering
  \centerline{\includegraphics[width=4.5cm]{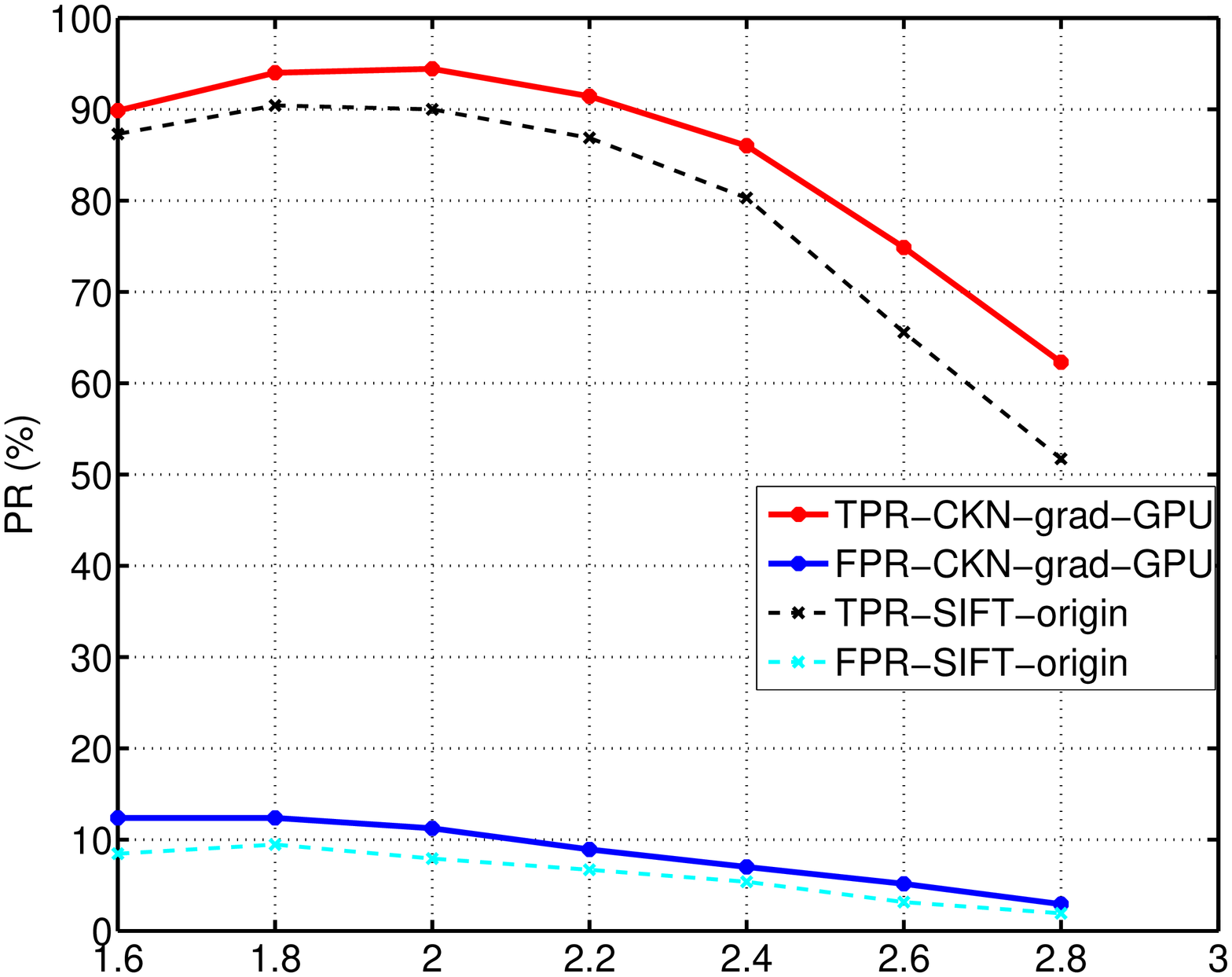}}
  \centerline{(b)single}
\end{minipage}
\hfill
\begin{minipage}[b]{0.3\linewidth}
  \centering
  \centerline{\includegraphics[width=4.5cm]{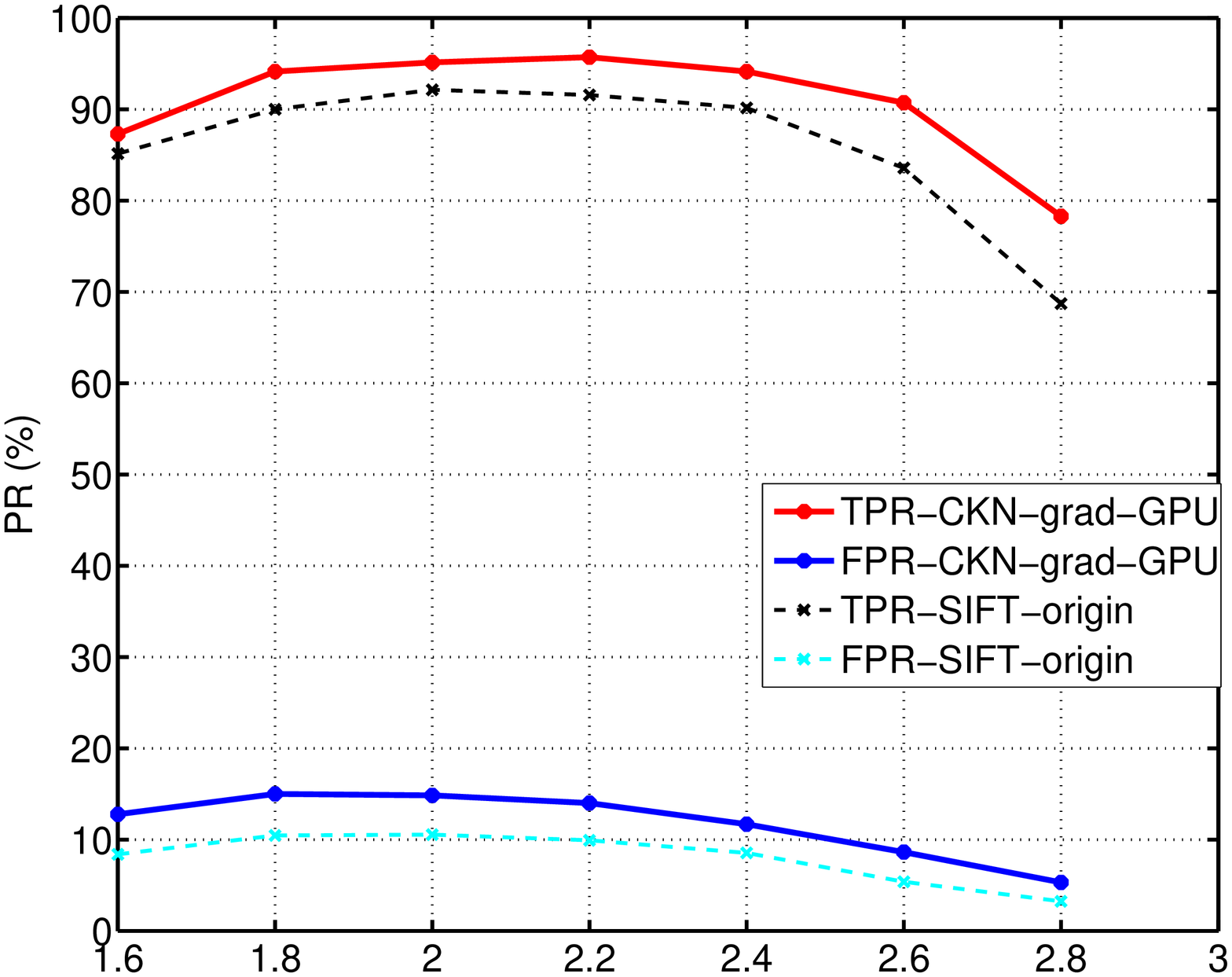}}
  \centerline{(c)centroid}
\end{minipage}
  \caption{The comparison between SIFT-origin and CKN-grad-GPU on MICC-F220 (top row) and MICC-F2000 (bottom row) for different linkage metrics and $T_h$ (axis x).}
  \label{fig:origin}
\end{figure*}

\begin{figure*}[htp]
\begin{minipage}[b]{0.3\linewidth}
  \centering
  \centerline{\includegraphics[width=4.5cm]{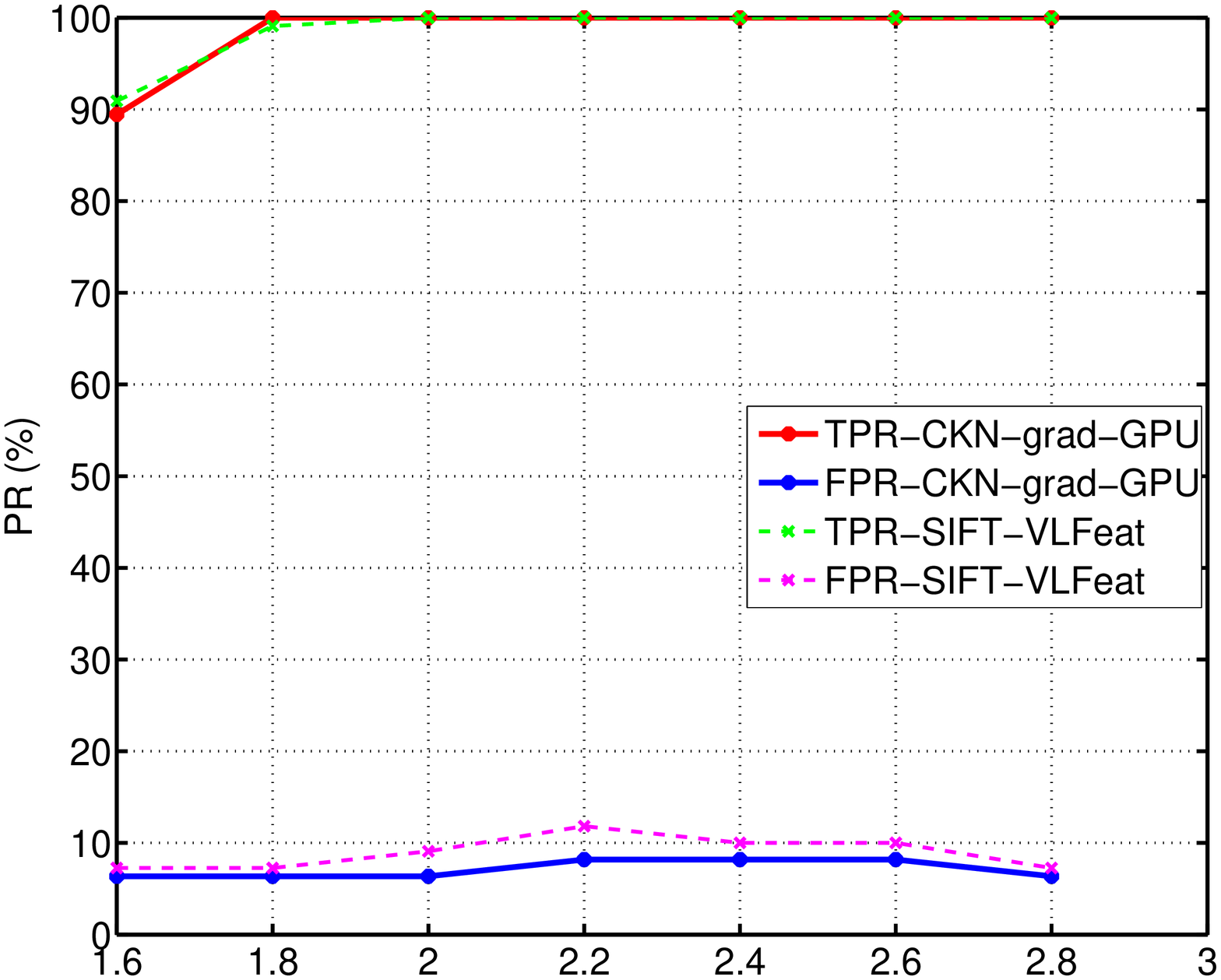}}
  \centerline{ }
\end{minipage}
\hfill
\begin{minipage}[b]{0.3\linewidth}
  \centering
  \centerline{\includegraphics[width=4.5cm]{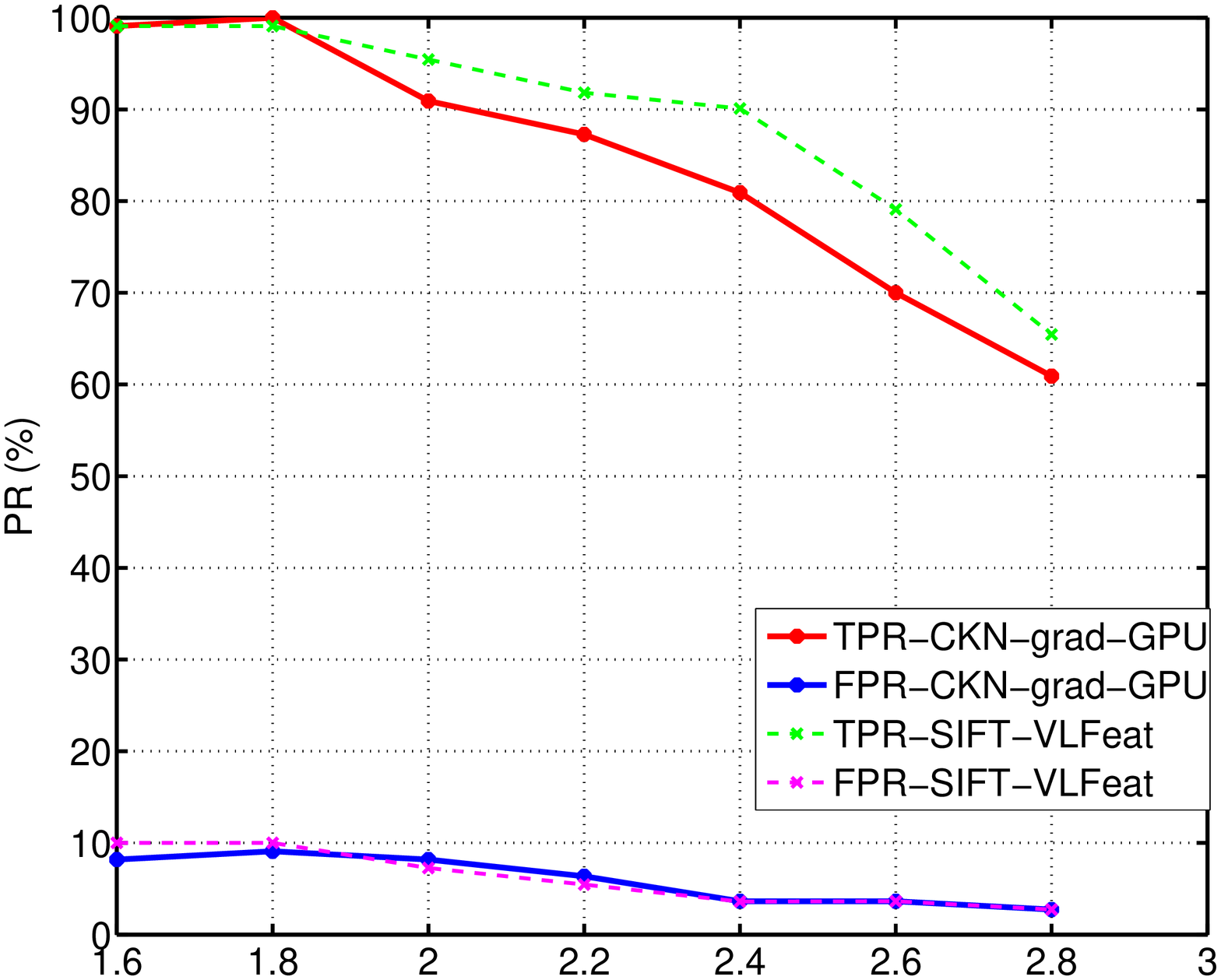}}
  \centerline{ }
\end{minipage}
\hfill
\begin{minipage}[b]{0.3\linewidth}
  \centering
  \centerline{\includegraphics[width=4.5cm]{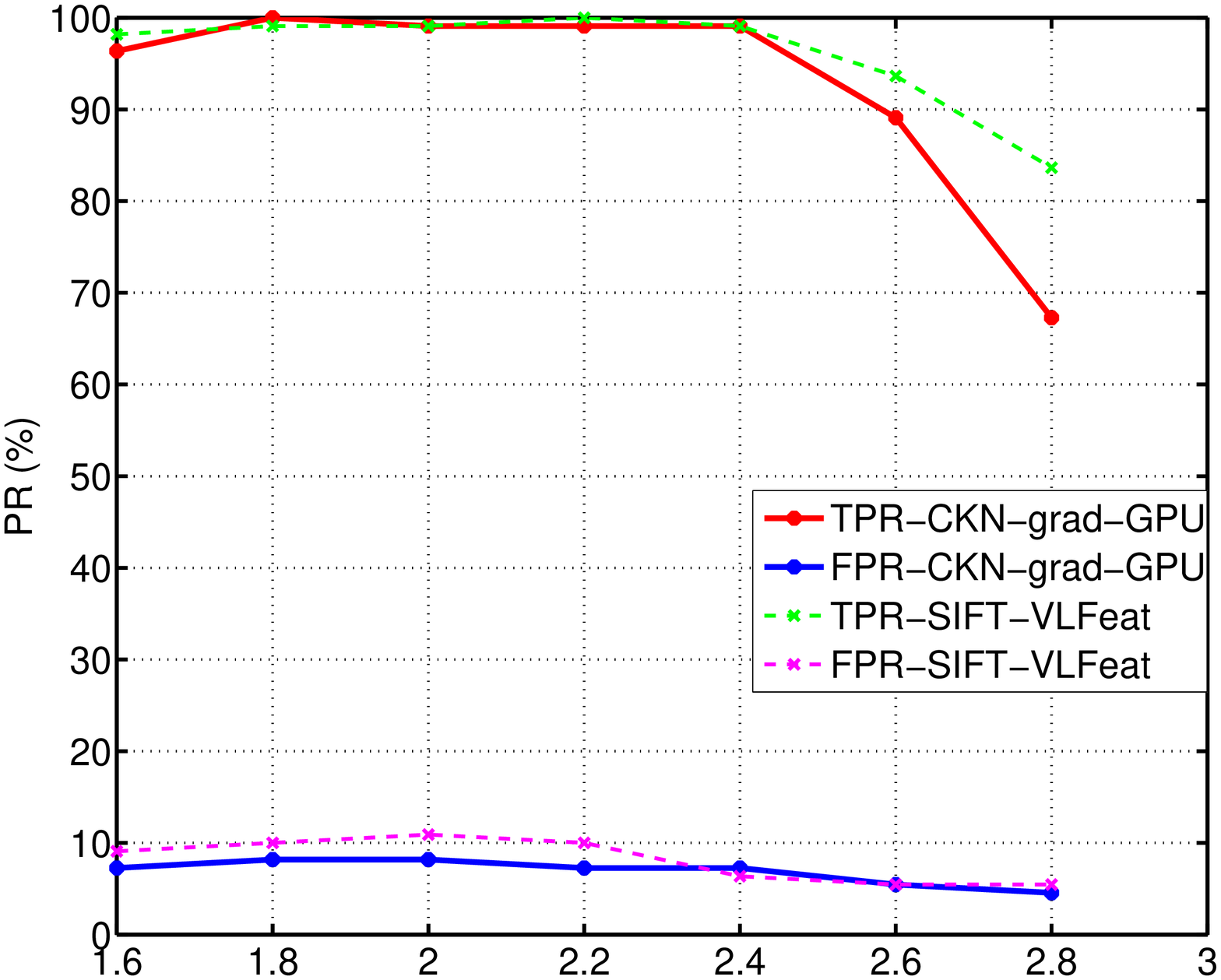}}
  \centerline{ }
\end{minipage}
\vfill
\begin{minipage}[b]{0.3\linewidth}
  \centering
  \centerline{\includegraphics[width=4.5cm]{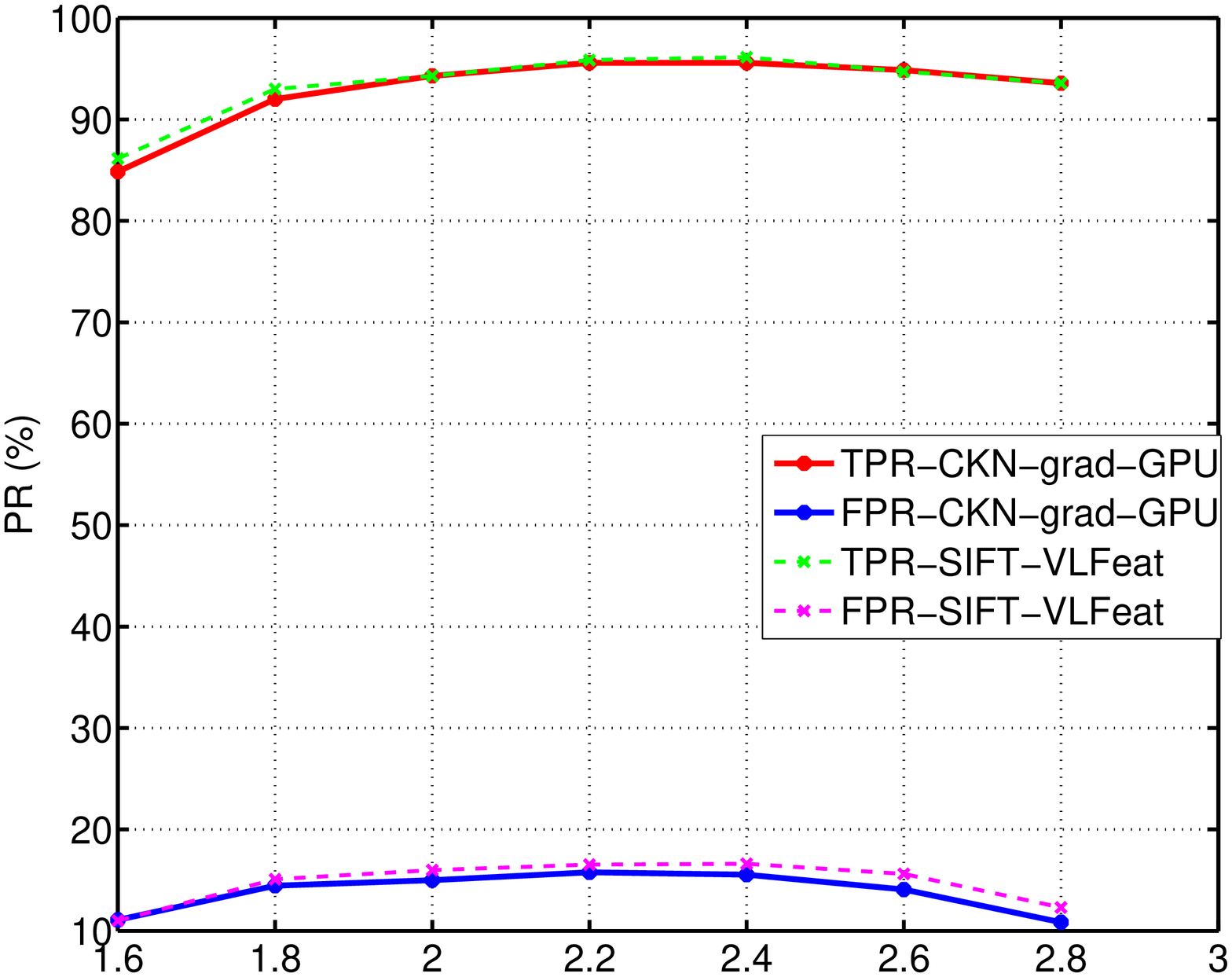}}
  \centerline{(a)ward}
\end{minipage}
\hfill
\begin{minipage}[b]{0.3\linewidth}
  \centering
  \centerline{\includegraphics[width=4.5cm]{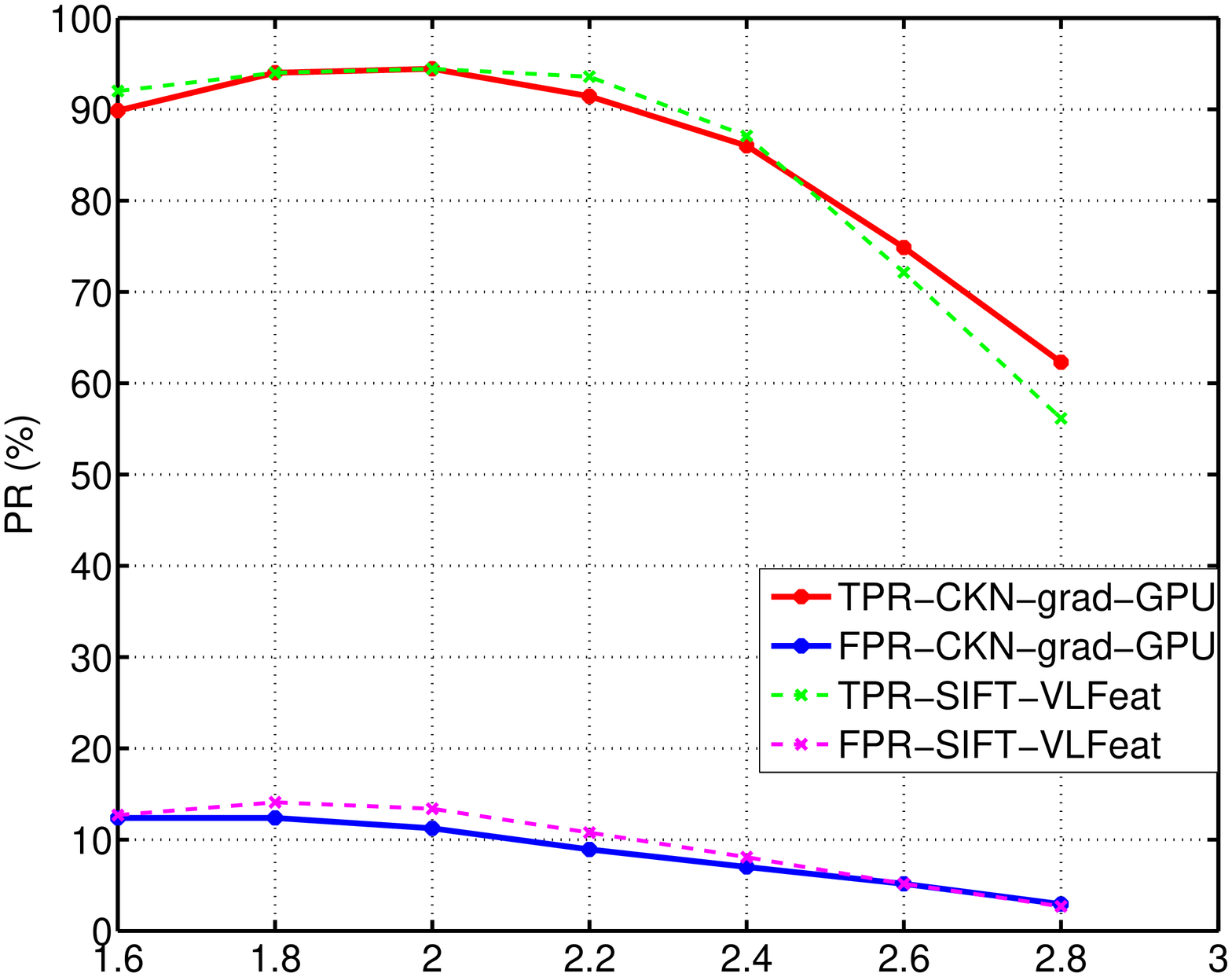}}
  \centerline{(b)single}
\end{minipage}
\hfill
\begin{minipage}[b]{0.3\linewidth}
  \centering
  \centerline{\includegraphics[width=4.5cm]{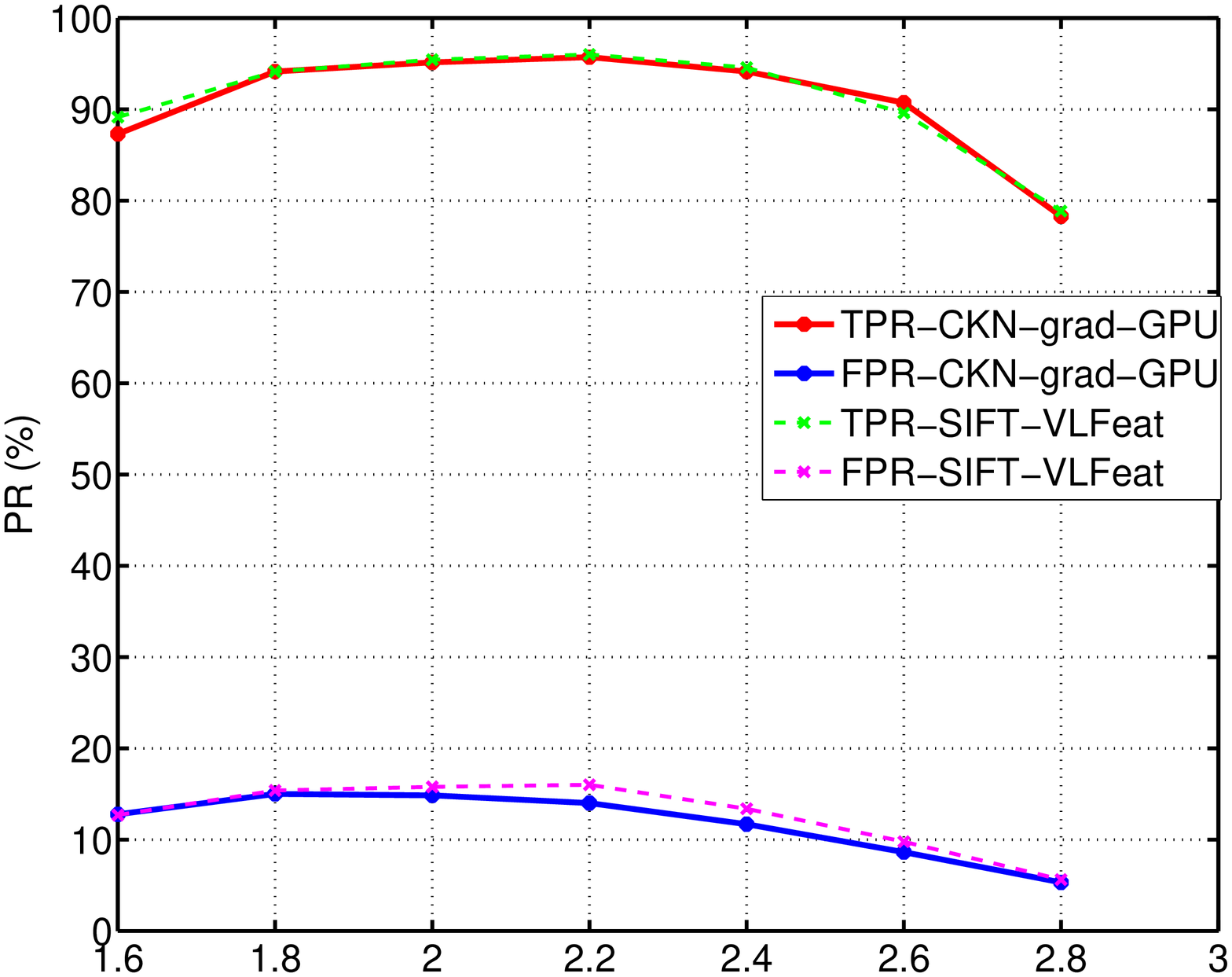}}
  \centerline{(c)centroid}
\end{minipage}
  \caption{The comparison between SIFT-VLFeat and CKN-grad-GPU on MICC-F220 (top row) and MICC-F2000 (bottom row) for different linkage metrics and $T_h$ (axis x).}
  \label{fig:vlfeat}
\end{figure*}

In Table \ref{table:F220}, the $T_h$ is set fixedly the same as the default parameter of provided codes while the linkage metric is varied. In this part, for each linkage method, we report the TPR and the FPR with respect to $T_h$, which varies in the interval $[1.6, 2.8]$ with steps of $0.2$. In general, \emph{CKN-grad-GPU} can achieve better performance than \emph{SIFT-origin} on both MICC-F220 and MICC-F2000, as shown in Fig. \ref{fig:origin}. Though the TPRs of \emph{CKN-grad-GPU} with the metric as single are lower while the $T_h$ is between $[2.0,2.8]$, \emph{CKN-grad-GPU} achieves the best performance while the $T_h$ is set as $1.6$ and $1.8$. On MICC-2000, \emph{CKN-grad-GPU} constantly achieves higher TPRs, but the FPRs become higher. As for the comparison between \emph{SIFT-VLFeat} and \emph{CKN-grad-GPU}, they achieve similar TPRs, while \emph{CKN-grad-GPU} achieves lower FPRs for most, as shown in Fig. \ref{fig:vlfeat}. It can be demonstrated that the discriminative capability of \emph{CKN-grad-GPU} is better.

\subsection{Comparison with other methods}
\label{sect:COM}

In this section, the proposed method is compared with two state-of-the-art copy-move forgery detection methods, the one is proposed by Li et al. \cite{TIFs2015JLi} and the other is proposed by Silva et al. \cite{JVCIR2015ESilva}. The comparisons are made on the publicly available dataset named CoMoFoD \cite{tralic2013comofod}. CoMoFoD consists of $200$ tampered examples with the size of $512\times512$. Images are forged by copying a part of an original image and pasting it on a different location in the same image. Five types of transformations are applied, namely translation, rotation, scaling, distortion and combination. Each type of transformation contains $40$ examples. Besides the version with no postprocessing, the images are processed under $6$ kinds of postprocessing respectively, namely JPEG compression, noise adding, image blurring, brightness change, color reduction, and contrast adjustments. Thus, we conduct $7$ separate experiments, as shown in Table \ref{table:CoMoFoDorigin}-\ref{table:CoMoFoDCA3}. On the dataset without postprocessing, we also conduct experiments to demonstrate the necessity and feasibility of each step, as shown in Table \ref{table:CoMoFoDstep}. The results of Li et al. \cite{TIFs2015JLi} and Silva et al. \cite{JVCIR2015ESilva} are generated by the codes provided by the authors. Because the CoMoFod dataset provides the pixel-level groundtruth for each tampered image, we compute the pixel level precision, recall and F1-measure for each detected result as follows:
\begin {equation}\label{eq:precision}
precision=\frac{|\{CMF\ pixels\}\cap\{retrieved\ pixels\}|}{|\{retrieved\ pixels\}|}
\end {equation}
\begin {equation}\label{eq:recall}
recall=\frac{|\{CMF\ pixels\}\cap\{retrieved\ pixels\}|}{|\{CMF\ pixels\}|}
\end {equation}
\begin {equation}\label{eq:F1}
F_1=2\times\frac{precision\times recall}{precision+recall}
\end {equation}
where $CMF\ pixels$ denotes the labeled tampered pixels in the ground truth, and $retrieved\ pixels$ denotes the detected tampered pixels. The precision calculates the ratio of the retrieved CMF pixels in all the retrieved pixels, and the recall calculates the ratio of the retrieved CMF pixels in all of the CMF pixels. For each generated binary tamper map, we computes its precision, recall and F1-measure, and then for different kinds of transformations, we computes their average precisions, recalls and F1-measures.

\begin{table*}[!t]
\renewcommand{\arraystretch}{1.3}
\caption{The performance of each step on CoMoFoD dataset with no postprocessing.}
\label{table:CoMoFoDstep}
\centering
\scriptsize
\begin{tabular}{c c c c c }
\hline
Algorithm & \tabincell{c}{Images with F1-\\measure $>0.5$} & \tabincell{c}{Average\\ precision} & \tabincell{c}{Average\\ recall} & \tabincell{c}{Average \\F1-measure} \\
\hline
Li et al. \cite{TIFs2015JLi} & 102 & 0.5446 & 0.8504 & 0.5954 \\
\cite{TIFs2015JLi} with COB & 100 & 0.5690 & 0.8156 & 0.6133 \\
\cite{TIFs2015JLi} with SKPD & 99 & 0.5390 & 0.8327 & 0.5838 \\
\cite{TIFs2015JLi} with COB+SKPD & 91 & 0.5662 & 0.8040 & 0.6055 \\
\cite{TIFs2015JLi} with COB+CKN & 90 & 0.5536 & 0.8193 & 0.6054 \\
Ours & 97 & 0.5927 & 0.8220 & 0.6318   \\
\hline
\end{tabular}
\end{table*}

In Table \ref{table:CoMoFoDstep}, comparisons are conducted with different settings. We set the work of Li et al. \cite{TIFs2015JLi} as the baseline which can achieve the state-of-the-art performance. While we simply replace the segmentation method as COB, both the precision and F1-measure increase. However, if we simply adopt the SKPD (segmentation-based keypoint distribution), all the scores decrease. If we adopt COB and SKPD simultaneously, the scores are lower than the version with COB. It seems like that SKPD is useless. However, we find that if we adopt the pipeline of \cite{TIFs2015JLi} with COB+CKN, the performance is even worse. Dramatically, with the help of SKPD (the full pipeline shown in Fig. \ref{fig:frameworkour}), our method can achieve good performance with the highest precision and F1-measure scores. The main reason is that SKPD provides more redundant keypoints and the discriminative capability of SIFT is not good enough, so the false alarmed areas are too much. While CKN is so cautious that many tampered areas are miss-detected. The redundant keypoints of SKPD can provide more candidates for CKN, and the combination of both can achieve better performance.

The comparisons with other methods on the images without further postprocessing are shown in Table \ref{table:CoMoFoDorigin}. It can be seen that for different kinds transformations, the proposed method can get higher average precisions than Li et al. \cite{TIFs2015JLi}, while the F1-measures are higher except for scaling transformation. Of course, Silva et al. \cite{JVCIR2015ESilva} can achieve excellent performance in some ways, e.g., the best performance for translation. In general, the proposed method can achieve better performance than the original work of Li et al. \cite{TIFs2015JLi} and Silva et al. \cite{JVCIR2015ESilva} on CoMoFoD dataset with no postprocessing.

Then we test the robustness of the proposed method under different attacks, namely, JPEG compression, noise adding, image blurring, brightness change, color reduction, and contrast adjustments. As shown in Table \ref{table:CoMoFoDorigin}-\ref{table:CoMoFoDCA3}, for most cases, the proposed method can get higher precisions and competitive F1 scores, and the proposed method is quite robust to different attacks. As for noise adding and brightness change, the number of images with F1-measure $>0.5$ decreases while the images are forged with scaling and combination transformation. In the tampered images with combination transformations, there are $27$ images (total $40$ images) have been under scaling transformation. Thus, the proposed method still needs to reinforce its scale invariant property. Overall, the proposed method can achieve competitive performance than the state-of-the-art methods, and is robust to different kinds of attacks.

\begin{table*}[!t]
\renewcommand{\arraystretch}{1.3}
\caption{Copy-move forgery detection results on CoMoFoD dataset with no postprocessing.}
\label{table:CoMoFoDorigin}
\centering
\scriptsize
\begin{tabular}{c c c c c c }
\hline
 Algorithm & \tabincell{c}{Transformation\\(number of images)} & \tabincell{c}{Images with F1-\\measure $>0.5$} & \tabincell{c}{Average\\precision} & \tabincell{c}{Average\\recall} & \tabincell{c}{Average \\ F1-measure} \\
\hline
\multirow{5}{*}{\tabincell{c}{Li et al.\\ \cite{TIFs2015JLi}}} & Translation (40) & 17 & 0.4180 & 0.8327 & 0.4798   \\
 & Rotation(40) & 19 & 0.5594 & 0.8281 & 0.5978 \\
  & Scaling(40) & 20 & 0.5542 & 0.8492 & 0.6059 \\
   & Distortion(40) & 26 & 0.6425 & 0.9045 & 0.6961 \\
    & Combination(40) & 20 & 0.5448 & 0.8306 & 0.5919 \\
\hline
\multirow{5}{*}{\tabincell{c}{Silva et al.\\ \cite{JVCIR2015ESilva}}} & Translation (40) & 20 & 0.4921 & 0.7754 & 0.5493  \\
 & Rotation(40) & 17 & 0.5532 & 0.7451 & 0.5573 \\
  & Scaling(40) & 14 & 0.4966 & 0.6971 & 0.5008 \\
   & Distortion(40) & 18 & 0.5814 & 0.7878 & 0.5650 \\
    & Combination(40) & 19 & 0.5885 & 0.6934 & 0.5475 \\
\hline
\multirow{5}{*}{Ours} & Translation (40) & 16 & 0.4547 & 0.8023 & 0.5246 \\
 & Rotation(40) & 24 & 0.6833 & 0.9006 & 0.7174 \\
  & Scaling(40) & 16 & 0.5696 & 0.7516 & 0.5864 \\
   & Distortion(40) & 25 & 0.6631 & 0.8516 & 0.6987 \\
    & Combination(40) & 16 & 0.5599 & 0.7825 & 0.5997 \\
\hline
\end{tabular}
\end{table*}

\begin{table*}[!t]
\renewcommand{\arraystretch}{1.3}
\caption{Copy-move forgery detection results on CoMoFoD dataset with JPEG compression (quality factor = 90).}
\label{table:CoMoFoDJC8}
\centering
\scriptsize
\begin{tabular}{c c c c c c }
\hline
 Algorithm & \tabincell{c}{Transformation\\(number of images)} & \tabincell{c}{Images with F1-\\measure $>0.5$} & \tabincell{c}{Average\\precision} & \tabincell{c}{Average\\recall} & \tabincell{c}{Average \\ F1-measure} \\
\hline
\multirow{5}{*}{\tabincell{c}{Li et al.\\ \cite{TIFs2015JLi}}} & Translation (40) & 13 & 0.3835 & 0.8473 & 0.4502 \\
 & Rotation(40) & 16 & 0.5809 & 0.8817 & 0.6285 \\
  & Scaling(40) & 20 & 0.5630 & 0.8448 & 0.6037 \\
   & Distortion(40) & 22 & 0.6490 & 0.8860 & 0.6902 \\
    & Combination(40) & 18 & 0.5528 & 0.8816 & 0.5879 \\
\hline
\multirow{5}{*}{\tabincell{c}{Silva et al.\\ \cite{JVCIR2015ESilva}}} & Translation (40) & 7 & 0.3789 & 0.4122 & 0.3113  \\
 & Rotation(40) & 4 & 0.3990 & 0.3779 & 0.2708 \\
  & Scaling(40) & 4 & 0.4858 & 0.3567 & 0.3000 \\
   & Distortion(40) & 7 & 0.5625 & 0.3825 & 0.3571 \\
    & Combination(40) & 9 & 0.5355 & 0.4170 & 0.3447 \\
\hline
\multirow{5}{*}{Ours} & Translation (40) & 12 & 0.4052 & 0.7260 & 0.4658 \\
 & Rotation(40) & 16 & 0.5977 & 0.8014 & 0.6369 \\
  & Scaling(40) & 15 & 0.5169 & 0.7248 & 0.5449 \\
   & Distortion(40) & 22 & 0.5963 & 0.7787 & 0.6175 \\
    & Combination(40) & 13 & 0.5162 & 0.7260 & 0.5138 \\
\hline
\end{tabular}
\end{table*}

\begin{table*}[!t]
\renewcommand{\arraystretch}{1.3}
\caption{Copy-move forgery detection results on CoMoFoD dataset with Noise adding (variance = 0.0005).}
\label{table:CoMoFoDNA3}
\centering
\scriptsize
\begin{tabular}{c c c c c c }
\hline
 Algorithm & \tabincell{c}{Transformation\\(number of images)} & \tabincell{c}{Images with F1-\\measure $>0.5$} & \tabincell{c}{Average\\precision} & \tabincell{c}{Average\\recall} & \tabincell{c}{Average \\ F1-measure} \\
\hline
\multirow{5}{*}{\tabincell{c}{Li et al.\\ \cite{TIFs2015JLi}}} & Translation (40) & 13 & 0.4636 & 0.7563 & 0.5211 \\
 & Rotation(40) & 16 & 0.6202 & 0.8399 & 0.6528 \\
  & Scaling(40) & 18 & 0.5673 & 0.7438 & 0.5849 \\
   & Distortion(40) & 20 & 0.6806 & 0.7821 & 0.7013 \\
    & Combination(40) & 16 & 0.5466 & 0.7411 & 0.5575 \\
\hline
\multirow{5}{*}{\tabincell{c}{Silva et al.\\ \cite{JVCIR2015ESilva}}} & Translation (40) & 5 & 0.4035 & 0.4550 & 0.3170  \\
 & Rotation(40) & 8 & 0.5924 & 0.6481 & 0.5265 \\
  & Scaling(40) & 9 & 0.6159 & 0.5115 & 0.4987 \\
   & Distortion(40) & 14 & 0.6828 & 0.5627 & 0.5270 \\
    & Combination(40) & 10 & 0.5998 & 0.4912 & 0.4681 \\
\hline
\multirow{5}{*}{Ours} & Translation (40) & 14 & 0.5097 & 0.7819 & 0.5623 \\
 & Rotation(40) & 15 & 0.6385 & 0.8076 & 0.6578 \\
  & Scaling(40) & 9 & 0.5838 & 0.6840 & 0.5677 \\
   & Distortion(40) & 19 & 0.7380 & 0.8411 & 0.7627 \\
    & Combination(40) & 10 & 0.5842 & 0.7705 & 0.6086 \\
\hline
\end{tabular}
\end{table*}

\begin{table*}[!t]
\renewcommand{\arraystretch}{1.3}
\caption{Copy-move forgery detection results on CoMoFoD dataset with image blurring (averaging filter $= 3\times3$).}
\label{table:CoMoFoDIB1}
\centering
\scriptsize
\begin{tabular}{c c c c c c }
\hline
 Algorithm & \tabincell{c}{Transformation\\(number of images)} & \tabincell{c}{Images with F1-\\measure $>0.5$} & \tabincell{c}{Average\\precision} & \tabincell{c}{Average\\recall} & \tabincell{c}{Average \\ F1-measure} \\
\hline
\multirow{5}{*}{\tabincell{c}{Li et al.\\ \cite{TIFs2015JLi}}} & Translation (40) & 13 & 0.3186 & 0.9206 & 0.4067 \\
 & Rotation(40) & 18 & 0.4481 & 0.8753 & 0.5280 \\
  & Scaling(40) & 18 & 0.4514 & 0.9096 & 0.5304 \\
   & Distortion(40) & 24 & 0.5022 & 0.9449 & 0.5953 \\
    & Combination(40) & 18 & 0.4139 & 0.9008 & 0.4961 \\
\hline
\multirow{5}{*}{\tabincell{c}{Silva et al.\\ \cite{JVCIR2015ESilva}}} & Translation (40) & 19 & 0.4842 & 0.7653 & 0.5356  \\
 & Rotation(40) & 17 & 0.5183 & 0.7043 & 0.5335 \\
  & Scaling(40) & 17 & 0.5281 & 0.6994 & 0.5212 \\
   & Distortion(40) & 22 & 0.6243 & 0.8292 & 0.6048 \\
    & Combination(40) & 19 & 0.5383 & 0.6873 & 0.5281 \\
\hline
\multirow{5}{*}{Ours} & Translation (40) & 14 & 0.3481 & 0.8270 & 0.4318 \\
 & Rotation(40) & 22 & 0.5114 & 0.8591 & 0.5945 \\
  & Scaling(40) & 18 & 0.4890 & 0.7836 & 0.5540 \\
   & Distortion(40) & 29 & 0.5715 & 0.8949 & 0.6611 \\
    & Combination(40) & 21 & 0.4849 & 0.8448 & 0.5575 \\
\hline
\end{tabular}
\end{table*}

\begin{table*}[!t]
\renewcommand{\arraystretch}{1.3}
\caption{Copy-move forgery detection results on CoMoFoD dataset with brightness change ((lower bound, upper bound) $= (0.01,0.8)$).}
\label{table:CoMoFoDBC3}
\centering
\scriptsize
\begin{tabular}{c c c c c c }
\hline
 Algorithm & \tabincell{c}{Transformation\\(number of images)} & \tabincell{c}{Images with F1-\\measure $>0.5$} & \tabincell{c}{Average\\precision} & \tabincell{c}{Average\\recall} & \tabincell{c}{Average \\ F1-measure} \\
\hline
\multirow{5}{*}{\tabincell{c}{Li et al.\\ \cite{TIFs2015JLi}}} & Translation (40) & 15 & 0.3957 & 0.7942 & 0.4623 \\
 & Rotation(40) & 16 & 0.5601 & 0.8445 & 0.5933 \\
  & Scaling(40) & 22 & 0.5537 & 0.7860 & 0.5926 \\
   & Distortion(40) & 23 & 0.6464 & 0.8964 & 0.6892 \\
    & Combination(40) & 14 & 0.5446 & 0.8857 & 0.5889 \\
\hline
\multirow{5}{*}{\tabincell{c}{Silva et al.\\ \cite{JVCIR2015ESilva}}} & Translation (40) & 16 & 0.4157 & 0.7429 & 0.4775  \\
 & Rotation(40) & 16 & 0.5272 & 0.7314 & 0.5333 \\
  & Scaling(40) & 14 & 0.3977 & 0.6305 & 0.4378 \\
   & Distortion(40) & 18 & 0.4834 & 0.7168 & 0.5137 \\
    & Combination(40) & 18 & 0.5583 & 0.6733 & 0.5225 \\
\hline
\multirow{5}{*}{Ours} & Translation (40) & 14 & 0.4128 & 0.7848 & 0.4773 \\
 & Rotation(40) & 15 & 0.6075 & 0.9063 & 0.6531 \\
  & Scaling(40) & 10 & 0.5350 & 0.7290 & 0.5526 \\
   & Distortion(40) & 20 & 0.6342 & 0.7814 & 0.6609 \\
    & Combination(40) & 13 & 0.5410 & 0.8468 & 0.5902 \\
\hline
\end{tabular}
\end{table*}

\begin{table*}[!t]
\renewcommand{\arraystretch}{1.3}
\caption{Copy-move forgery detection results on CoMoFoD dataset with color reduction (intensity levels per each color channel $= 32$).}
\label{table:CoMoFoDCR1}
\centering
\scriptsize
\begin{tabular}{c c c c c c }
\hline
 Algorithm & \tabincell{c}{Transformation\\(number of images)} & \tabincell{c}{Images with F1-\\measure $>0.5$} & \tabincell{c}{Average\\precision} & \tabincell{c}{Average\\recall} & \tabincell{c}{Average \\ F1-measure} \\
\hline
\multirow{5}{*}{\tabincell{c}{Li et al.\\ \cite{TIFs2015JLi}}} & Translation (40) & 15 & 0.3940 & 0.8361 & 0.4698 \\
 & Rotation(40) & 19 & 0.5692 & 0.8340 & 0.6174 \\
  & Scaling(40) & 21 & 0.5628 & 0.8969 & 0.6251 \\
   & Distortion(40) & 30 & 0.6352 & 0.9226 & 0.6955 \\
    & Combination(40) & 22 & 0.5665 & 0.8772 & 0.6115 \\
\hline
\multirow{5}{*}{\tabincell{c}{Silva et al.\\ \cite{JVCIR2015ESilva}}} & Translation (40) & 19 & 0.4872 & 0.7884 & 0.5440  \\
 & Rotation(40) & 16 & 0.5432 & 0.6567 & 0.5066 \\
  & Scaling(40) & 14 & 0.5004 & 0.7092 & 0.5129 \\
   & Distortion(40) & 21 & 0.6147 & 0.7584 & 0.5813 \\
    & Combination(40) & 18 & 0.6456 & 0.6852 & 0.5605 \\
\hline
\multirow{5}{*}{Ours} & Translation (40) & 15 & 0.4502 & 0.8448 & 0.5203 \\
 & Rotation(40) & 19 & 0.6583 & 0.8415 & 0.6891 \\
  & Scaling(40) & 18 & 0.5984 & 0.7832 & 0.6267 \\
   & Distortion(40) & 24 & 0.6652 & 0.8193 & 0.6827 \\
    & Combination(40) & 16 & 0.6341 & 0.8744 & 0.6685 \\
\hline
\end{tabular}
\end{table*}

\begin{table*}[!t]
\renewcommand{\arraystretch}{1.3}
\caption{Copy-move forgery detection results on CoMoFoD dataset with contrast adjustments ((lower bound, upper bound) $= (0.01,0.8)$).}
\label{table:CoMoFoDCA3}
\centering
\scriptsize
\begin{tabular}{c c c c c c }
\hline
 Algorithm & \tabincell{c}{Transformation\\(number of images)} & \tabincell{c}{Images with F1-\\measure $>0.5$} & \tabincell{c}{Average\\precision} & \tabincell{c}{Average\\recall} & \tabincell{c}{Average \\ F1-measure} \\
\hline
\multirow{5}{*}{\tabincell{c}{Li et al.\\ \cite{TIFs2015JLi}}} & Translation (40) & 14 & 0.4031 & 0.8706 & 0.4746 \\
 & Rotation(40) & 18 & 0.5447 & 0.8428 & 0.5972 \\
  & Scaling(40) & 20 & 0.5714 & 0.8400 & 0.5973 \\
   & Distortion(40) & 27 & 0.6445 & 0.8994 & 0.6941 \\
    & Combination(40) & 20 & 0.5564 & 0.8586 & 0.5995 \\
\hline
\multirow{5}{*}{\tabincell{c}{Silva et al.\\ \cite{JVCIR2015ESilva}}} & Translation (40) & 18 & 0.5156 & 0.7382 & 0.5518  \\
 & Rotation(40) & 16 & 0.5722 & 0.6987 & 0.5321 \\
  & Scaling(40) & 15 & 0.5334 & 0.7228 & 0.5059 \\
   & Distortion(40) & 24 & 0.6450 & 0.7854 & 0.6080 \\
    & Combination(40) & 17 & 0.5616 & 0.6642 & 0.5158 \\
\hline
\multirow{5}{*}{Ours} & Translation (40) & 13 & 0.4101 & 0.7218 & 0.4670 \\
 & Rotation(40) & 20 & 0.6157 & 0.8675 & 0.6590 \\
  & Scaling(40) & 16 & 0.5517 & 0.7987 & 0.5948 \\
   & Distortion(40) & 28 & 0.6610 & 0.8476 & 0.6924 \\
    & Combination(40) & 17 & 0.5740 & 0.7854 & 0.6053 \\
\hline
\end{tabular}
\end{table*}

\section{Conclusion}
\label{sect:Conclusion}

In this paper, we propose a copy-move forgery detection method based on Convolutional Kernel Network, and reformulate Convolutional Kernel Network based on GPU. The main contributions can be concluded as follows: CKN adoption in copy-move forgery detection and GPU-based CKN reformulation, segmentation-based keypoints distribution strategy and GPU-based adaptive oversegmentation. Extensive experiments are conducted to show that the proposed method based on Convolutional Kernel Network can achieve competitive performance than conventional hand-crafted features and the state-of-the-art methods. By bridging a gap between copy-move forgery detection and data-driven local convolutional features, we believe that we are opening a fruitful research direction for the future.

\begin{acknowledgements}
This work was supported by the NSFC under U1636102 and U1536105, and National Key Technology R\&D Program under 2014BAH41B01, 2016YFB0801003 and 2016QY15Z2500.
\end{acknowledgements}


\bibliographystyle{spmpsci}
\bibliography{sigproc}

%
%

\end{document}